%% file: main.tex
\documentclass[runningheads]{llncs}

\usepackage{eccv}

\usepackage{eccvabbrv}

\input{preamble}
\usepackage[accsupp]{axessibility}  

\begin{document}

\title{GraphVid: Interactive Graph-Controllable \\Video Generation} 

\titlerunning{GraphVid: Interactive Graph-Controllable Video Generation}

\author{ Vedant Shah\inst{1} \and Onkar Susladkar\inst{1} \and Tushar Prakash\inst{2} \and Kiet A. Nguyen\inst{1} \and Tianjiao Yu\inst{1} \and Adheesh Juvekar\inst{1} \and Muntasir Wahed\inst{1} \and Ismini Lourentzou\inst{1} } \authorrunning{V. Shah et al.} \institute{ \begin{tabular}{c@{\hspace{2cm}}c} $^{1}${University of Illinois Urbana-Champaign} & $^{2}$Sony Research India \\[0.5em] \multicolumn{2}{c}{\email{\{vrshah4,onkarks2,lourent2\}@illinois.edu}} \end{tabular} }

\input{assets_tex/figures}

\maketitle
\FigAbstract

\input{assets_tex/tables}

\input{sections/0_abstract}   

\input{sections/1_introduction}

\input{sections/2_related_work}
\input{sections/3_methodology}
\input{sections/4_dataset}
\input{sections/5_experiments}

\input{sections/6_conculsion}


%
%
\bibliographystyle{splncs04}
\bibliography{main}

\input{sections/7_supplementary}
\end{document}

%% file: preamble.tex
\usepackage{hyperref}

\hypersetup{hypertexnames=false}
\usepackage{enumitem}
\usepackage{graphicx}
\usepackage{booktabs}
\usepackage[table]{xcolor}
\usepackage{colortbl}
\definecolor{best}{RGB}{255,210,170}      
\definecolor{second}{RGB}{255,235,210}    
\definecolor{grayrow}{RGB}{235,235,235}

\usepackage{tcolorbox}
\tcbuselibrary{breakable}

\usepackage{orcidlink}
\newcommand{\modelnamenc}{\textbf{GraphVid}\xspace}
\newcommand{\modelname}{{GraphVid}}

\definecolor{IllinoisOrange}{HTML}{FF5F05}
\definecolor{IllinoisBlue}{HTML}{13294B}
\newcommand{\logoicon}{%
  \raisebox{-0.20\height}{\includegraphics[height=1.2em]{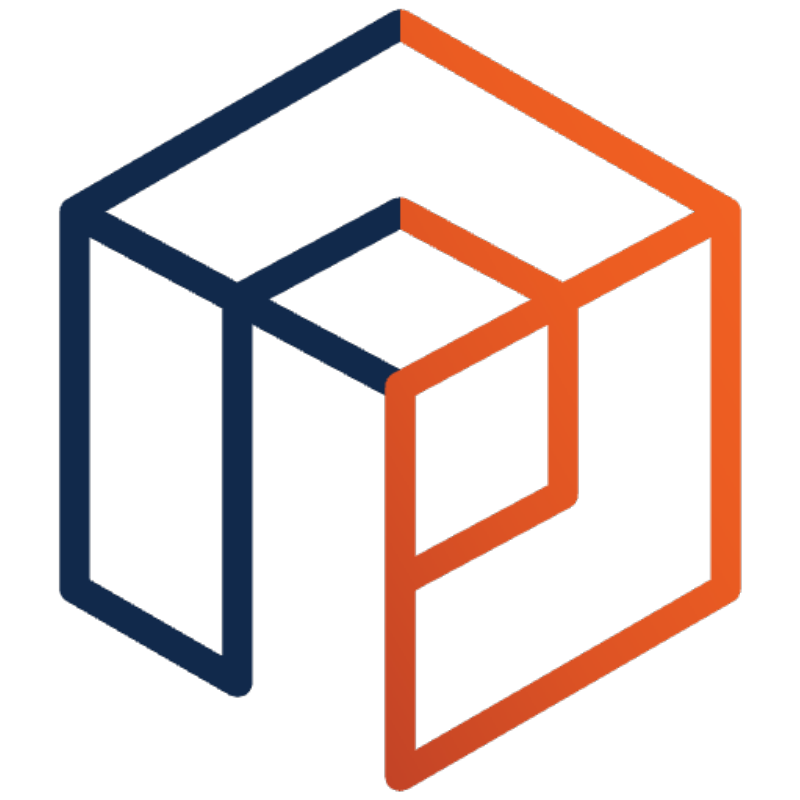}}%
}

\usepackage{pifont}
\usepackage{amsmath}
\usepackage{amssymb}
\usepackage[table,x11names]{xcolor}
\usepackage{tabulary}
\usepackage{colortbl}
\usepackage{booktabs}
\usepackage[font=footnotesize,skip=3pt,subrefformat=parens]{subcaption}

\usepackage{xfp}
\usepackage{epsfig}
\usepackage{graphicx}
\usepackage{tikz}
\usepackage{pgfplots}
\pgfplotsset{compat=1.18}
\usepackage{pgfplotstable}
\usepackage{caption}
\usepgfplotslibrary{groupplots}
\usetikzlibrary{patterns}

\usepackage{listings}
\usepackage{xcolor}
\usepackage{wrapfig}
\usepackage{tcolorbox}

\lstdefinestyle{custom}{
    backgroundcolor=\color{lightgray!20}, 
    basicstyle=\ttfamily\small,          
    columns=fullflexible,                
    frame=single,                        
    breaklines=true,                     
    postbreak=\mbox{\textcolor{red}{$\hookrightarrow$}\space}, 
}
\definecolor{LGray}{gray}{0.97}

\definecolor{SageGreen}{HTML}{e6f2ce}
\definecolor{DarkSageGreen}{HTML}{5d8412}
\definecolor{DarkRed}{HTML}{a01313}

\definecolor{BackgroundGreen}{HTML}{e6f2ce}
\definecolor{BackgroundPink}{HTML}{f4dcee}
\definecolor{BackgroundBlue}{HTML}{c3e8f4}
\definecolor{BackgroundCoffee}{HTML}{efdabf}
\definecolor{TextGreen}{HTML}{354c08}
\definecolor{TextPink}{HTML}{841c6e}
\definecolor{TextBlue}{HTML}{196577}
\definecolor{TextCoffee}{HTML}{684014}

\usepackage{xspace}
\usepackage{multirow}

\definecolor{LightCyan2}{RGB}{224, 255, 255}

\definecolor{lightgray}{rgb}{0.95, 0.95, 0.95}
\definecolor{darkgray}{rgb}{0.4, 0.4, 0.4}
\definecolor{purple}{rgb}{0.65, 0.12, 0.82}
\definecolor{mydarkgreen}{rgb}{0.0, 0.5, 0.0}  

\newtcolorbox{planbox}[1]{
    colback=gray!5,
    colframe=gray!75,
    title=#1,
    fonttitle=\bfseries
}

%% file: assets_tex/figures.tex
\newcommand{\FigAbstract}{
\begin{figure}[h]
\vspace{-0.8cm}
    \centering
    \includegraphics[width=.99\linewidth]{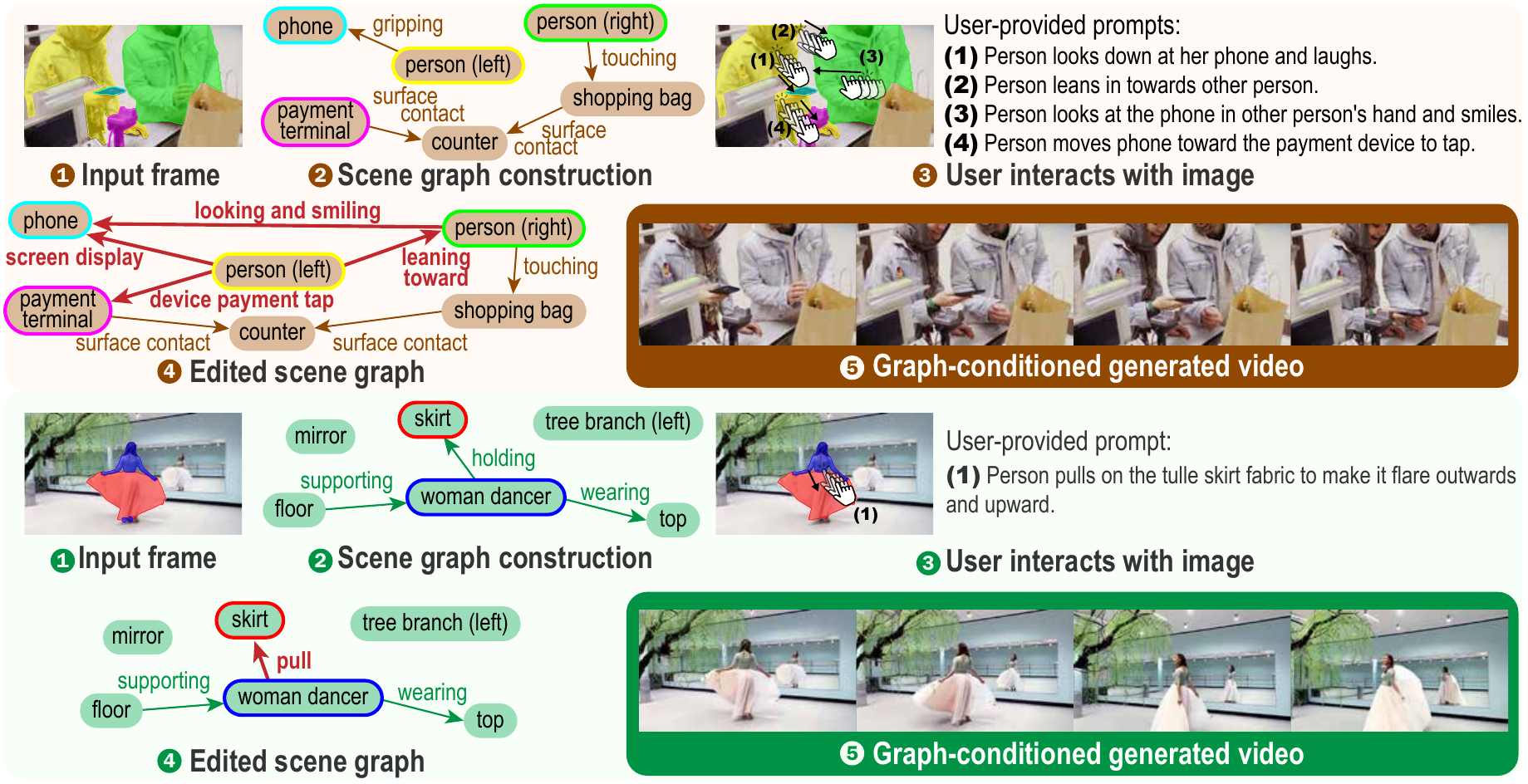}
    \vspace{-0.3cm}
        \caption{\textbf{\modelnamenc enables controllable multi-object image-to-video generation through interaction graphs.} Starting from an input frame, GraphVid constructs a scene graph that represents entities and their interactions. Users can directly modify these interactions by interacting with the image to specify desired dynamics. The edited interaction graph is then translated into conditioning tokens that guide a frozen video diffusion backbone to generate a coherent video sequence.}
    \label{fig:abstract_diagram}
    \vspace{-1cm}
\end{figure}
}

\newcommand{\FigMethod}{
\begin{figure*}[t]
    \centering
    \includegraphics[width=.99\linewidth]{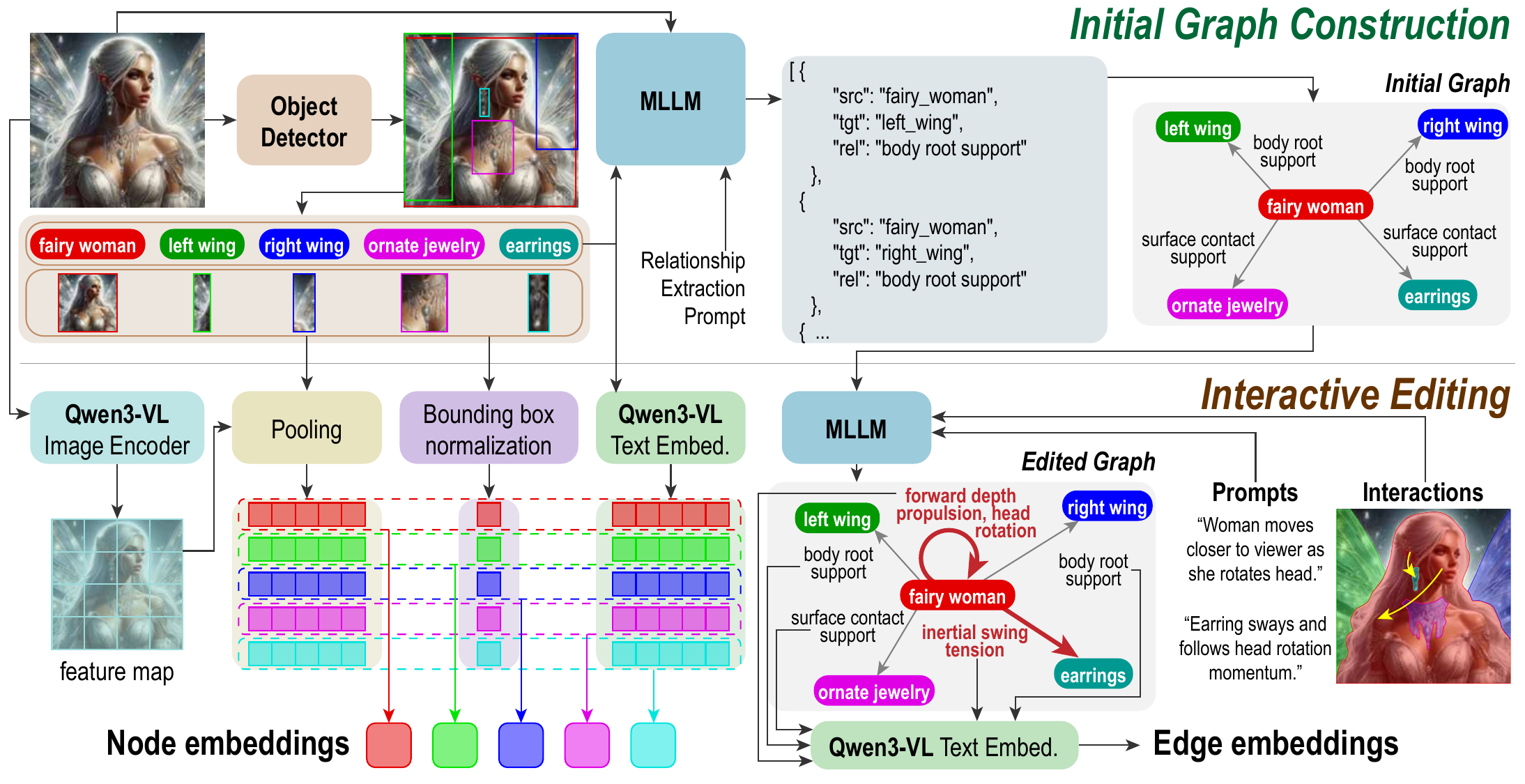}
    \vspace{-0.4cm}
\caption{\textbf{\modelnamenc Overview.} From an input image, objects are detected and encoded into node embeddings, while an MLLM extracts relational edges to construct an initial interaction graph. Edge-aware graph reasoning converts node and edge representations into conditioning tokens that guide a frozen video diffusion transformer to generate interaction-consistent video dynamics.}
\label{fig:method_arch_diagram}
\vspace{-0.3cm}
\end{figure*}
}

\newcommand{\FigMethodLtx}{
\begin{figure*}[t]
    \centering
    \includegraphics[width=.99\linewidth]{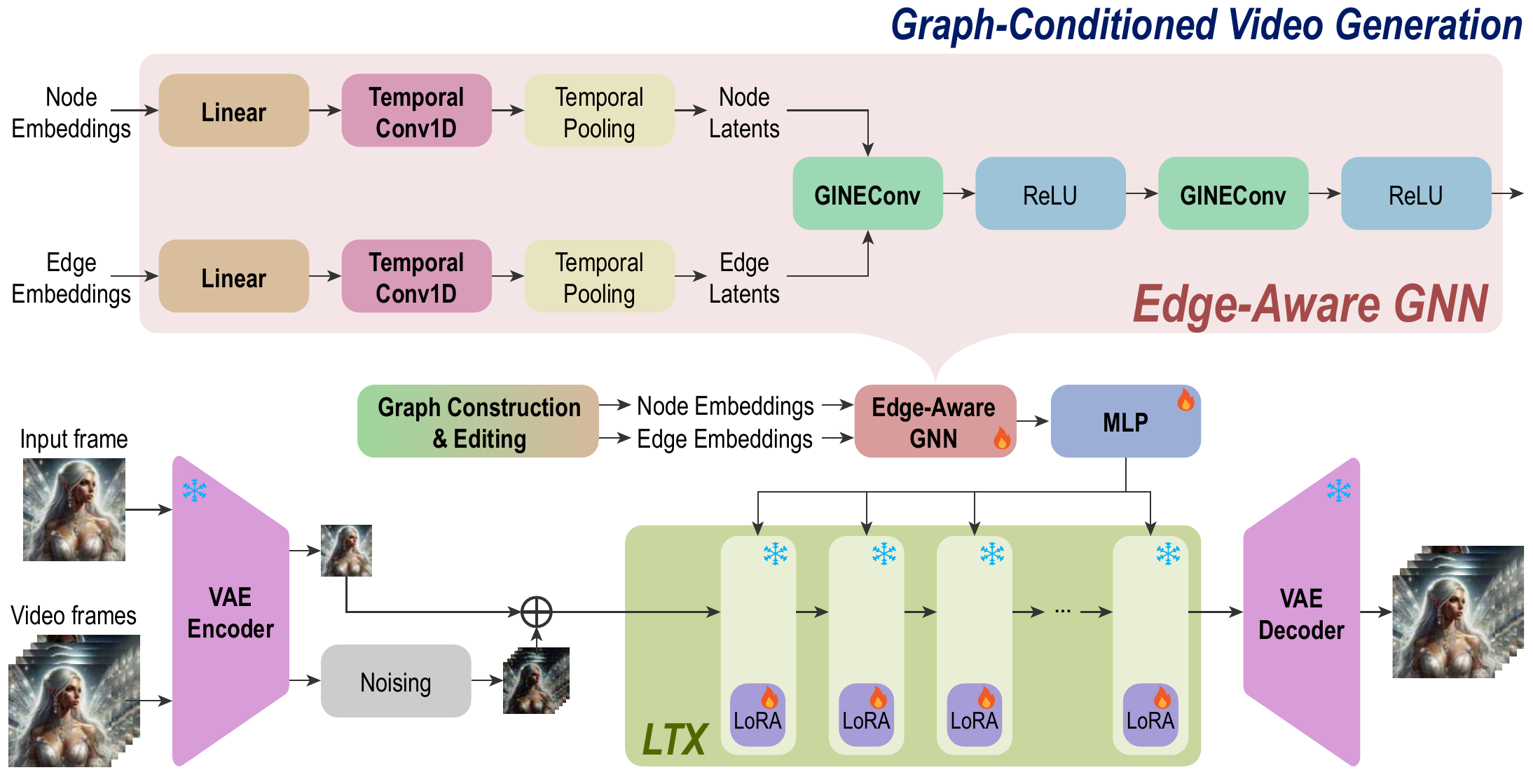}
    \vspace{-0.4cm}
\caption{\textbf{\modelnamenc Edge-Aware Graph Reasoning.} Node and edge embeddings are encoded into temporal latents and processed by our proposed edge-aware GNN that injects interaction semantics directly into message passing. The resulting graph embeddings capture relational dynamics between entities and are mapped into conditioning tokens that modulate a frozen diffusion video backbone via LoRA adapters.}
\label{fig:method_arch_ltx_diagram}
\vspace{-0.3cm}
\end{figure*}
}

\newcommand{\FigEfficiency}{
\begin{figure}[t]
    \centering
    \includegraphics[width=0.9\textwidth]{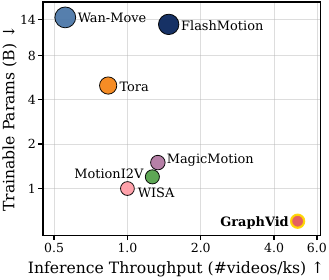}
    \caption{
    Efficiency comparison across video generation methods on distilled \textsc{MoveBench}.
    The x-axis shows inference throughput, while the y-axis indicates the number of 
    trainable parameters (in billions). Each marker corresponds to a method.
    \modelnamenc achieves competitive performance with significantly fewer
    trainable parameters while maintaining strong inference efficiency,
    highlighting the parameter efficiency of graph-based conditioning compared 
    to alternative approaches.
    }
    \label{fig:efficiency_tradeoff_movebench}
\end{figure}
}

\newcommand{\FigQualitative}{
\begin{figure*}[t]
\vspace{-0.2cm}
\centering
\includegraphics[width=0.98\textwidth]{assets_pdf/our_video_samples.pdf}
\vspace{-0.3cm}
\caption{
\textbf{Qualitative results of GraphVid for controllable interaction-based video generation.}
Given a single input image and user-specified interaction context, 
GraphVid generates temporally coherent video sequences reflecting the desired dynamics.
Examples include object-object interactions (geese moving relative to each other), human-object interactions (kicking a ball, holding a cup), and articulated motion (cycling).
The generated sequences demonstrate consistent motion, object preservation, and faithful execution of user-specified interaction dynamics.
}
\label{fig:graphvid_qualitative_results}
\vspace{-0.3cm}
\end{figure*}
}

\newcommand{\FigDatasetStats}{
\begin{figure}[t]
\centering
\begin{subfigure}[t]{0.53\linewidth}
\centering
\includegraphics[width=\linewidth]{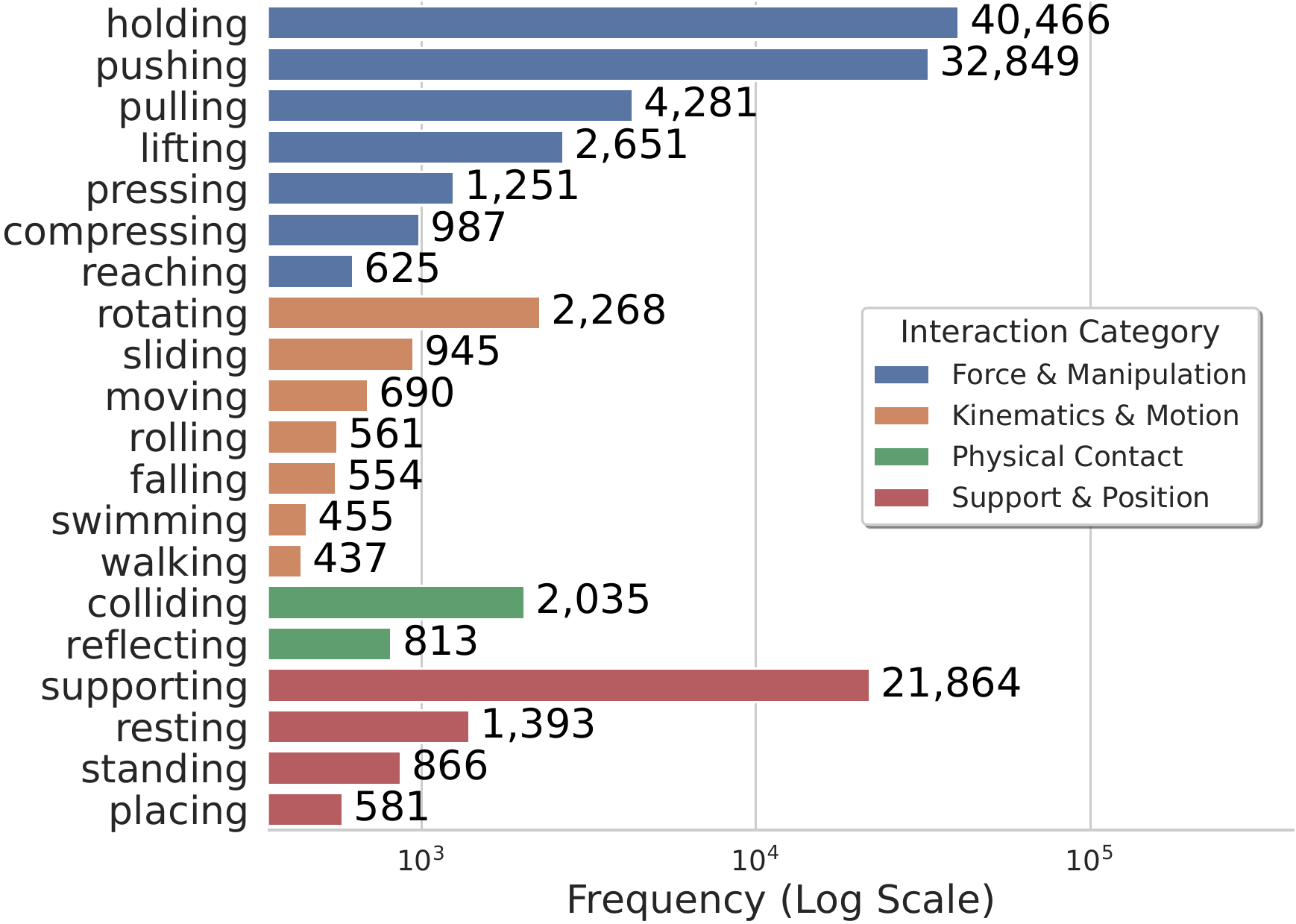}
\caption{Taxonomy of interaction types.}
\label{fig:interaction_taxonomy}
\end{subfigure}
\hfill
\begin{subfigure}[t]{0.46\linewidth}
\centering
\includegraphics[width=\linewidth]{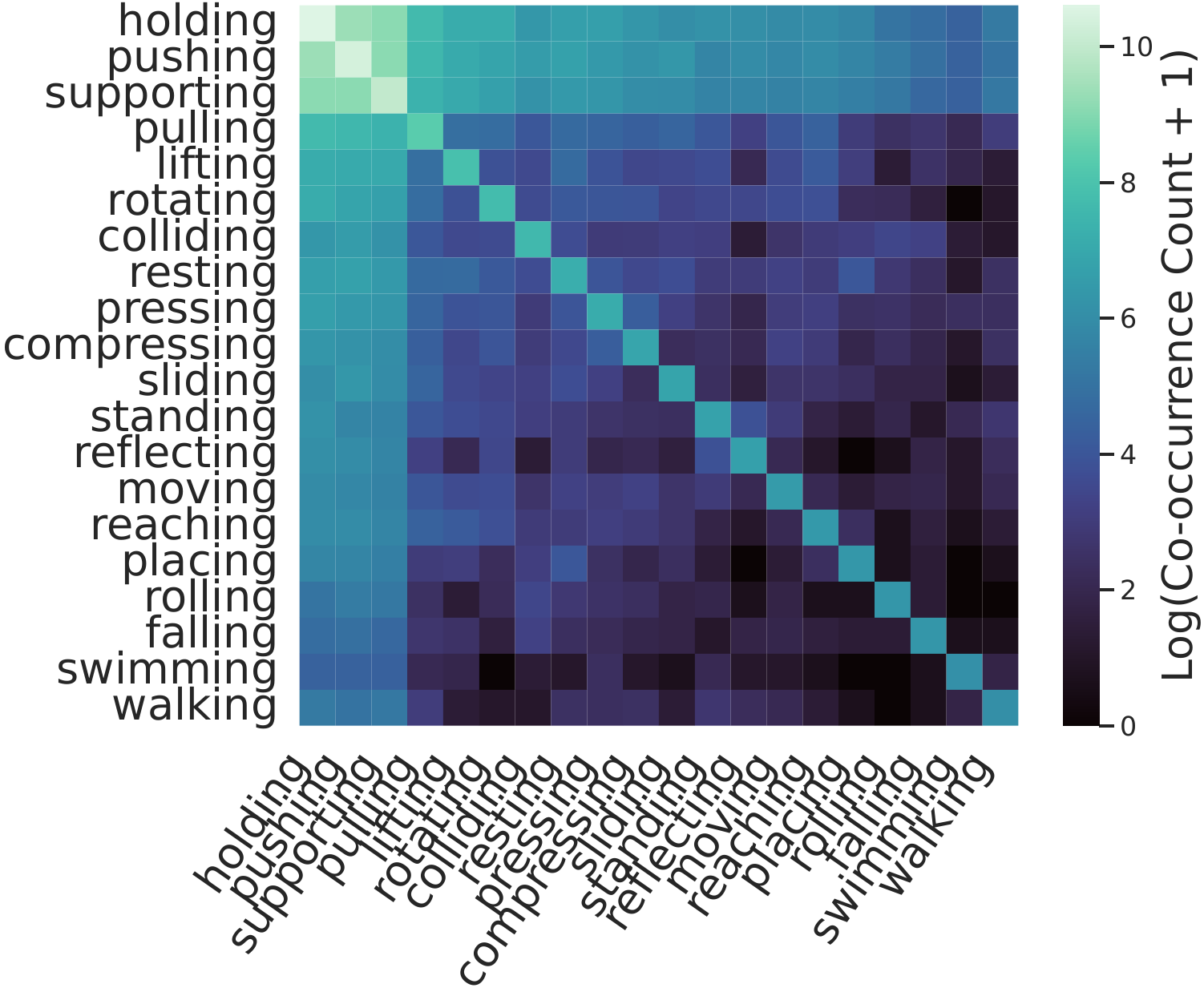}
\caption{Co-occurrence Statistics.}
\label{fig:interaction_cooccurrence}
\end{subfigure}
\vspace{-0.3cm}
\caption{
\textbf{Interaction statistics in \textsc{GraphVid-Bench} dataset.}
(a) Frequency of interaction primitives, grouped into four categories (force \& manipulation, kinematics \& motion, physical contact, and support \& position). 
(b) Interaction co-occurrence matrix captures interaction primitives appearing together within the same video sequences. The co-occurrence of manipulation and motion primitives shows that real-world dynamics are inherently compositional, motivating the use of structured interaction graphs for controllable video generation.
}
\label{fig:dataset_statistics}
\vspace{-0.3cm}
\end{figure}
}

\newcommand{\FigMoveBenchSubset}{
\begin{figure*}[t]
\centering
\includegraphics[width=0.99\textwidth]{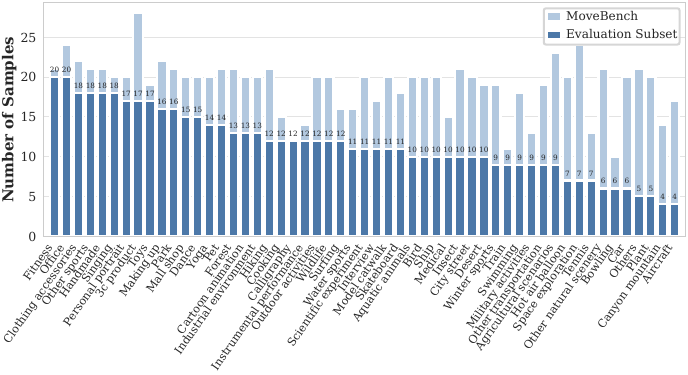}
\vspace{-0.4cm}
\caption{
\textbf{Category distribution of the \textsc{MoveBench} dataset and the curated evaluation subset.}
The figure shows the number of samples per semantic category in the original \textsc{MoveBench} dataset (light bars) and the subset selected for evaluation in our experiments (dark bars). 
Categories are ordered by decreasing subset frequency. 
Our curated evaluation split preserves broad semantic coverage while reducing extreme long-tail categories with few samples.
Importantly, the subset retains 53 out of 54 unique categories present in \textsc{MoveBench} ({98.15\% coverage}), ensuring that evaluation remains representative of the full dataset while focusing on interaction-rich scenarios.
}
\label{fig:movebench_subset_distribution}
\vspace{-0.2cm}
\end{figure*}
}

\newcommand{\FigSuppQualitativeComparisonSingle}{
\begin{figure*}[t]
\centering
\includegraphics[width=0.99\textwidth]{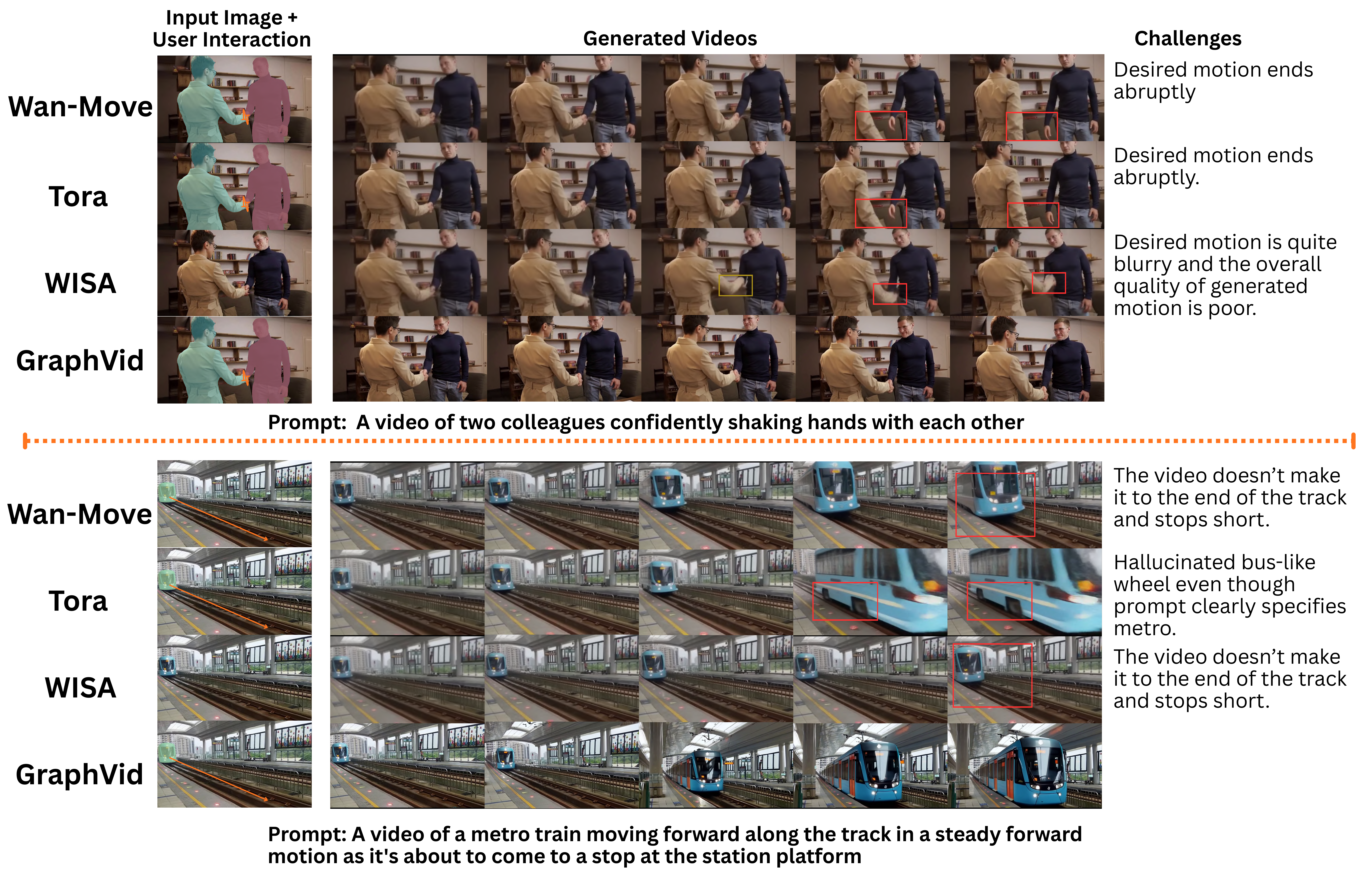}
\vspace{-0.35cm}
\caption{
\textbf{Qualitative comparison on interaction-driven video generation.}
Given an input image and user-specified interaction cues (left), different models generate video sequences over time (center). Highlighted regions mark common failure modes of prior methods. In the handshake scenario (top), baseline models exhibit abrupt motion termination or blurry hand dynamics, whereas GraphVid produces a consistent handshake with smooth temporal continuity. In the metro scenario (bottom), competing methods stop prematurely or hallucinate structural artifacts (\eg, bus-like wheels), while GraphVid preserves object structure and generates stable forward motion along the intended trajectory.
}
\label{fig:qualitative_single_comp_interactions}
\end{figure*}
}

\newcommand{\FigSuppQualitativeComparisonMulti}{
\begin{figure*}[t]
\centering
\includegraphics[width=0.99\textwidth]{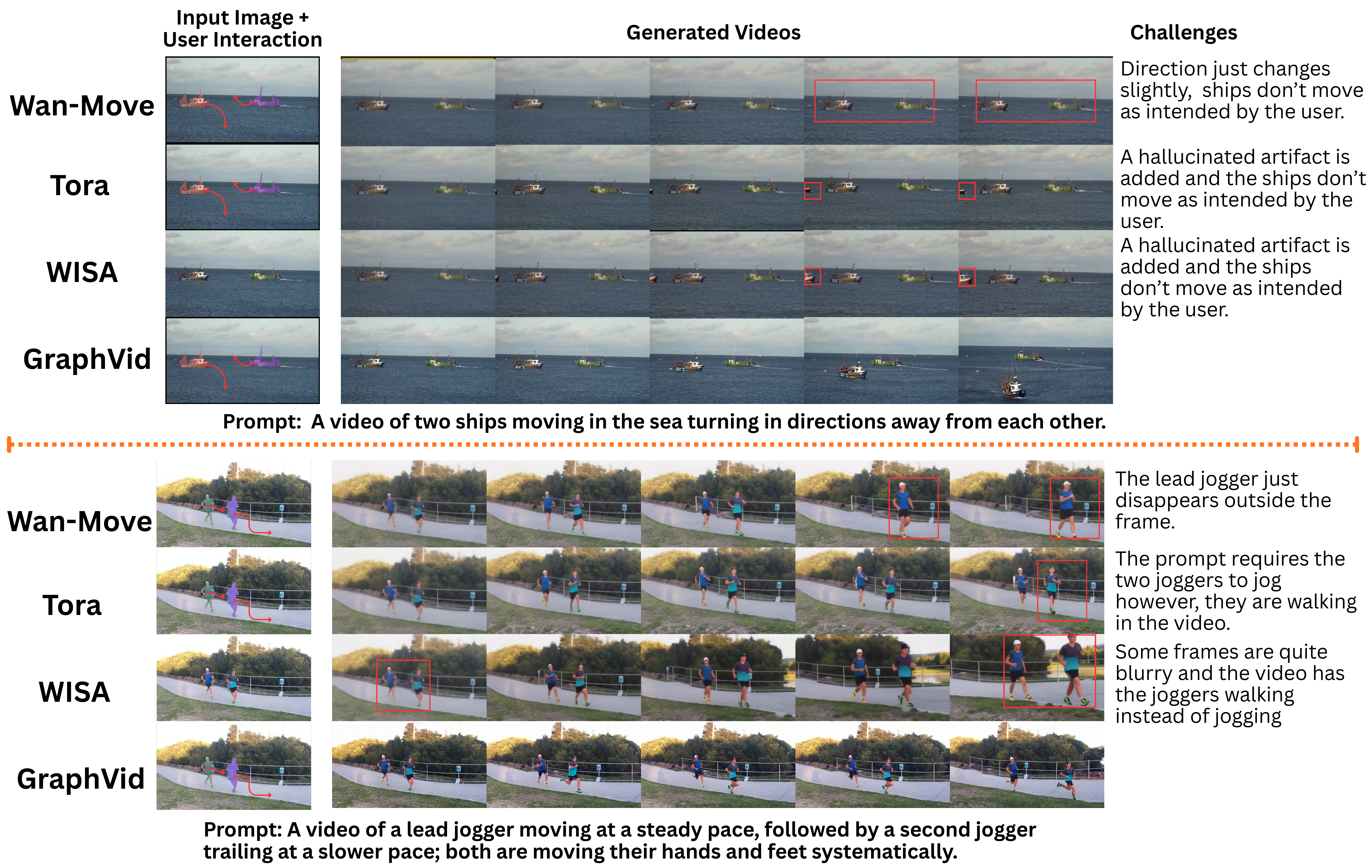}
\vspace{-0.35cm}
\caption{\textbf{Qualitative comparison on multi-object interaction control.} Given an input image and user-specified interaction cues (left), different methods generate video sequences over time (center). Highlighted regions indicate common failure modes in prior approaches. In the ship scenario (top), baseline methods struggle to follow the specified motion directions or introduce artifacts, whereas GraphVid produces coherent diverging trajectories consistent with the interaction cues. In the jogging scenario (bottom), competing methods show unstable motion (\eg, subject disappearance, walking instead of jogging, or blur), while GraphVid maintains stable identities and consistent multi-entity dynamics.}

\label{fig:qualitative_multi_comp_interactions}
\end{figure*}
}

\newcommand{\FigUserStudy}{
\begin{figure}[t!]
     \centering
     \includegraphics[width=0.99\linewidth]{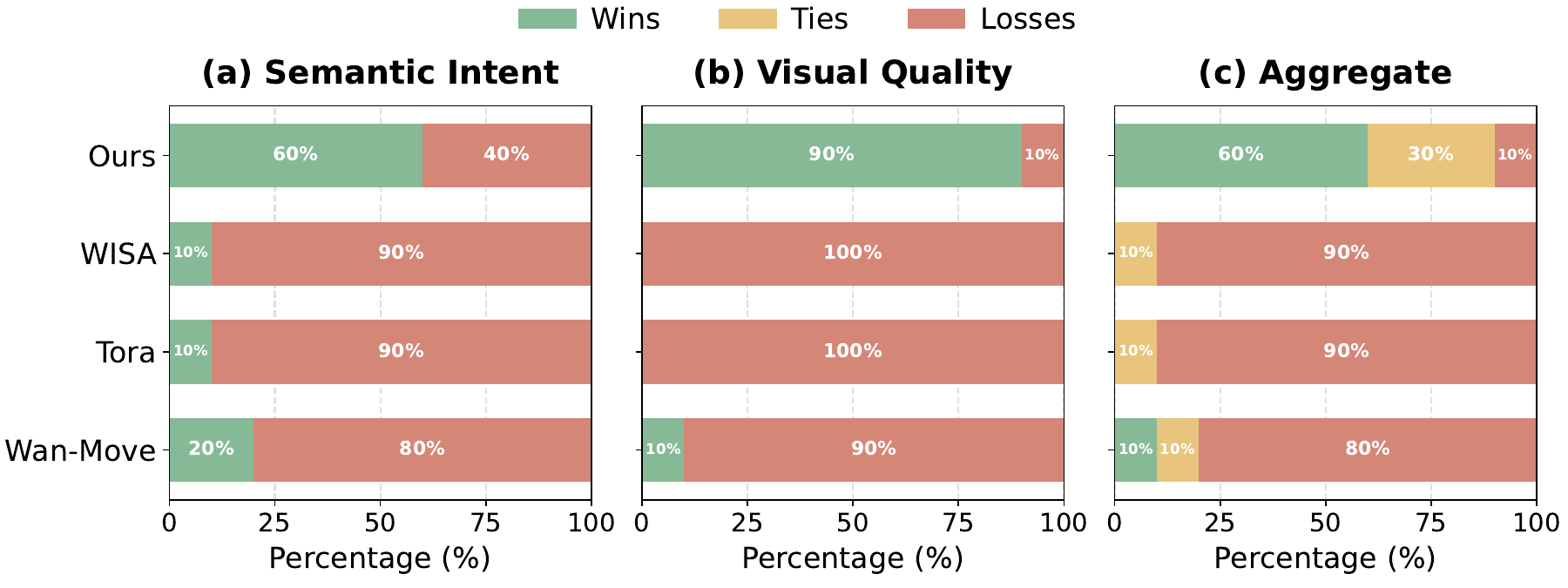}
     \vspace{-0.4cm}
     \caption{\textbf{User preference study across different dimensions}. GraphVid achieves the highest Selection Rate (Wins) compared to all baselines. Here, ``Wins'' denotes the percentage of times a model was chosen as the absolute best for a given metric, ``Ties'' indicates scenarios where multiple models were deemed equally preferred, and ``Losses'' indicates the model was not selected.}
     \label{fig:overall_comparison}
\end{figure}
}
\newcommand{\FigFailureZzz}{
\begin{figure}[t]
\vspace{-0.2cm}
\centering
\includegraphics[width=0.95\linewidth]{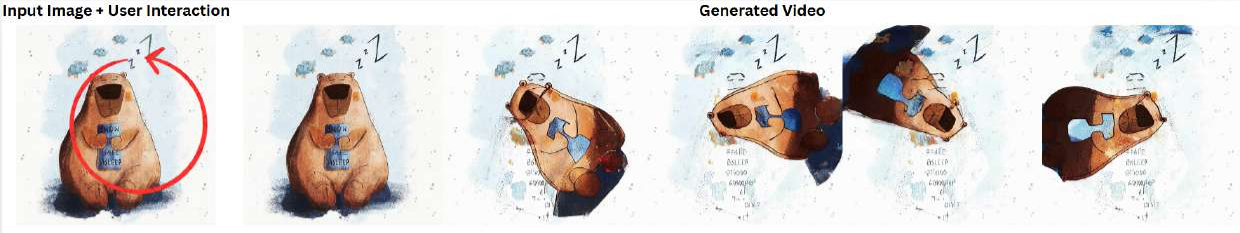}
\vspace{-0.3cm}
\caption{
\textbf{Failure case: semantic grounding error for small symbolic objects.}
The user interaction specifies a rotational transformation applied to the ``Zzz'' symbol, indicating sleep (left). 
However, the detector fails to isolate the text as an independent node during scene graph construction. 
Consequently, the motion transformation is incorrectly applied to the dominant object (the bear), producing severe geometric distortion and unrealistic deformation across frames. 
This example highlights a limitation of current vision-language detection when grounding small semantic elements that lack strong visual salience.
}
\label{fig:failure_semantic_granularity}
\end{figure}
}

\newcommand{\FigGeneratedSamples}{
\begin{figure*}[t]
\vspace{-1.2cm}
\centering
\includegraphics[width=0.98\textwidth]{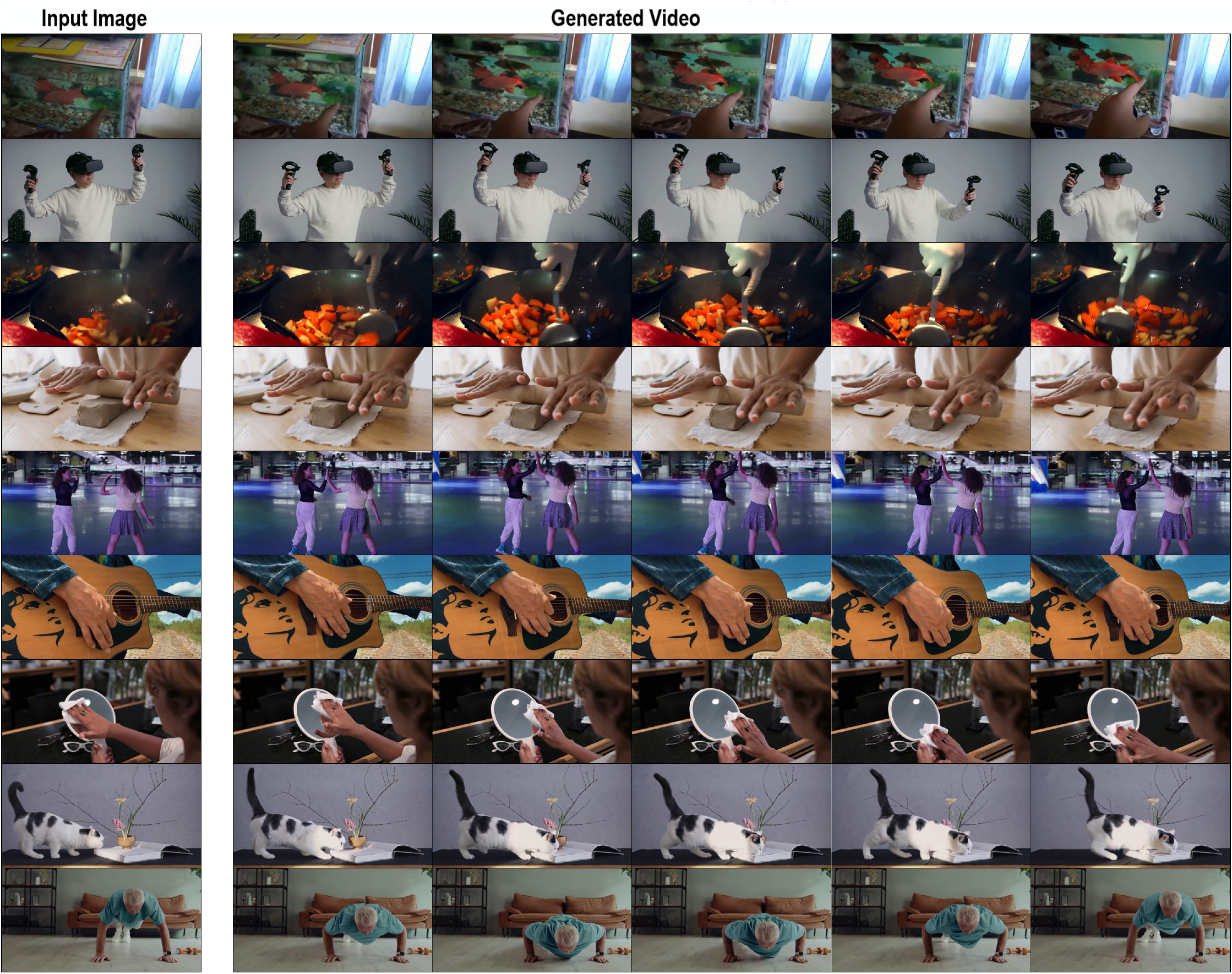}
\vspace{-0.3cm}
\caption{
\textbf{Additional qualitative results on diverse interaction scenarios.}
Each row shows a different example from our evaluation set.
The first column provides the input image, while the remaining columns show frames sampled from the generated video.
GraphVid produces temporally coherent motion across a wide range of activities, including object manipulation, human actions, and articulated motion such as cooking, sculpting, dancing, playing instruments, and animal movement.
These examples demonstrate that structured interaction conditioning enables consistent object dynamics while preserving scene appearance and identity across frames.
}
\label{fig:generated_video_samples}
\vspace{-0.4cm}
\end{figure*}
}

%% file: assets_tex/tables.tex
\newcommand{\TableMvAll}{\begin{table}[t!]
\centering
\caption{\textbf{Performance on Distilled \textsc{MoveBench}.}}
\label{tab:distilled-movebench}
\setlength{\tabcolsep}{4pt}
\renewcommand{\arraystretch}{1.05}
\resizebox{\linewidth}{!}{
\begin{tabular}{l|ccc|cccc}
\toprule
Method 
& \shortstack{Train\\Data} 
& \shortstack{Params\\(B)} 
& \shortstack{Inference\\(s)$\downarrow$}
& FID$\downarrow$ 
& FVD$\downarrow$ 
& PSNR$\uparrow$ 
& SSIM$\uparrow$ \\
\midrule
Wan-Move~\cite{wan2025wan} & 2M & 14.5 & 1800 & \textbf{15.56} & \textbf{82.17} & \textbf{17.21} & \textbf{0.61} \\
Tora~\cite{zhang2025tora} & 630k & 5 & 1200 & 24.45 & 110.47 & 11.27 & 0.54\\
Motion-I2V~\cite{shi2024motion} & 10M & 1.2 & 790 & 28.32 & 159.32 & 9.87 & 0.38 \\
MagicMotion~\cite{li2025magicmotion} & 23k & 1.5 & \underline{750} & 26.57 & \underline{105.12} & 12.04 & 0.51\\
WISA~\cite{wangwisa} & 80k & 1 & 1000 & 25.98 & 107.89 & 15.08 & 0.53\\
\textbf{Ours} & 27k & 0.3 & \textbf{520} & \underline{18.59} & 106.59 & \underline{15.79} & \underline{0.59}\\
\bottomrule
\end{tabular}}
\end{table}}

\newcommand{\TableMvMulti}{\begin{table}[t]
\centering
\caption{\textbf{Distilled \textsc{MoveBench} Multi-Object Motion Results.}}
\label{tab:distilled-movebench-multi}
\setlength{\tabcolsep}{5pt}
\renewcommand{\arraystretch}{1.1}

\begin{tabular}{l|cccc}
\toprule
\textbf{Method} 
& FID$\downarrow$ & FVD$\downarrow$ & PSNR$\uparrow$ & SSIM$\uparrow$ \\
\midrule
WISA~\cite{wangwisa}
& 60.12 & 341 & 13.13 & 0.55 \\
Tora~\cite{zhang2025tora}
& 56.04 & 369 & \underline{14.98} & 0.52 \\
Wan-Move~\cite{wan2025wan} 
& \textbf{31.29} & \textbf{252} & \textbf{16.69} & \textbf{0.61} \\
\textbf{Ours}
& \underline{54.12} & \underline{312} & 14.45 & \underline{0.55} \\

\bottomrule
\end{tabular}
\end{table}}

\newcommand{\TableAllAblations}{
\begin{table}[t]
\centering
\caption{\textbf{Ablation Studies.} Validating conditioning strategies, backbone generalization, and graph prompt density. All variants evaluated on Distilled \textsc{MoveBench}.}
\vspace{-2mm}
\label{tab:all-ablations}

\begin{minipage}[t]{0.56\textwidth}
\centering
{\small \textbf{(a) Conditioning Modalities} \label{tab:ablation-conditioning}} \\ \vspace{1mm}
\resizebox{\linewidth}{!}{
\begin{tabular}{l|cccc}
\toprule
Variant & FID$\downarrow$ & FVD$\downarrow$ & PSNR$\uparrow$ & SSIM$\uparrow$ \\
\midrule
Text + Img (Wan 2.2) & 23.42 & 117 & 12.88 & 0.56 \\
Graph (no edge text) & 21.92 & 114 & 13.23 & 0.57 \\
Graph (full) & \textbf{18.59} & \textbf{106.59} & \textbf{15.79} & \textbf{0.59} \\
\bottomrule
\end{tabular}}
\end{minipage}\hfill
\begin{minipage}[t]{0.42\textwidth}
\centering
{\small \textbf{(b) Backbone Generalization} \label{tab:ablation-backbone}} \\ \vspace{1mm}
\resizebox{\linewidth}{!}{
\begin{tabular}{l|cccc}
\toprule
Backbone & FID$\downarrow$ & FVD$\downarrow$ & PSNR$\uparrow$ & SSIM$\uparrow$ \\
\midrule
Ours + LTX & 18.72 & 105.88 & 15.42 & 0.58 \\
Ours + Wan 2.2 & \textbf{18.59} & \textbf{106.59} & \textbf{15.79} & \textbf{0.59} \\
\bottomrule
\end{tabular}}
\end{minipage}

\vspace{4mm} 

\begin{minipage}[t]{0.55\textwidth}
\centering
{\small \textbf{(c) Graph Prompt Density (Max Nodes)} \label{tab:ablation-graph-density}} \\ \vspace{1mm}
\resizebox{\linewidth}{!}{
\begin{tabular}{l|cccc}
\toprule
Max Nodes & FID$\downarrow$ & FVD$\downarrow$ & PSNR$\uparrow$ & SSIM$\uparrow$ \\
\midrule
50 & -- & -- & -- & -- \\
128 & \textbf{18.59} & \textbf{106.59} & \textbf{15.79} & \textbf{0.59} \\
256 & 19.09 & 108.12 & 15.12 & 0.58 \\
512 & 20.11 & 111.19 & 14.54 & 0.57 \\
\bottomrule
\end{tabular}}
\end{minipage}
\vspace{-3mm}
\end{table}
}

\newcommand{\TableDataStat}{
\begin{table}[t]
        \centering
        \caption{\textbf{Graph Dataset Statistics.} Across $\sim$27k curated videos. Interaction complexity demonstrates the prevalence of multi-entity dynamics.}
        \begin{tabular}{lccccc}
        \toprule
        \textbf{Metric} & \textbf{Mean} & \textbf{Median} & \textbf{Std. Dev.} & \textbf{Min} & \textbf{Max} \\
        \midrule
        Nodes per Graph & 7.76 & 5.0 & 8.03 & 1.0 & 78.0 \\
        Edges per Graph & 4.85 & 4.0 & 3.46 & 0.0 & 116.0 \\
        \midrule
        \multicolumn{6}{c}{\textbf{Interaction Complexity Breakdown}} \\
        \midrule
        \multicolumn{3}{l}{No-Interaction Graphs ($N=0$)} & \multicolumn{3}{r}{3881} \\
        \multicolumn{3}{l}{Single-Interaction Graphs ($N=1$)} & \multicolumn{3}{r}{10982} \\
        \multicolumn{3}{l}{Multi-Interaction Graphs ($N>1$)} & \multicolumn{3}{r}{12641} \\
        \bottomrule
        \end{tabular}
        \label{tab:dataset_stats}
\end{table}}

%% file: sections/0_abstract.tex
\begin{abstract}
Controllable video generation remains challenging due to the difficulty of specifying precise multi-object interactions using text prompts or motion-control inputs that primarily constrain pixel movement. In practice, trajectory-based control often requires users to draw accurate tracks for multiple objects, which scales poorly with scene complexity and becomes ambiguous under occlusion or overlap. To enable flexible yet precise multi-subject control, we introduce \modelnamenc, a graph-conditioned image-to-video generation model that enables interactive control through structured interaction graphs. 
We further curate \textsc{GraphVid-Bench}, a large-scale interaction-centric video dataset with structured relational annotations to enable training of interaction-aware video generation models. 
Despite using substantially less training data and fewer trainable parameters than prior motion-control methods, GraphVid delivers strong controllability and video quality. Compared with Motion-I2V, GraphVid reduces FID by up to 39.9\% and FVD by 37.6\%, while improving PSNR (9.87→15.98) and SSIM (0.38→0.61). Our results highlight the potential of structured semantic interfaces as a powerful paradigm for controllable video generation.\looseness-1 

\noindent \logoicon~\textcolor{IllinoisBlue}{PLAN Lab}~\href{https://plan-lab.github.io/graphvid}{\textcolor{IllinoisOrange}{https://plan-lab.github.io/graphvid}}

  \keywords{Controllable Video Generation \and Scene Graphs}
\end{abstract}

%% file: sections/1_introduction.tex
\section{Introduction}

Image-to-Video (I2V) generation has attracted increasing attention in generative modeling~\cite{blattmann2023stable, singer2022make, shi2024motion, li2025image, susladkar2025motionaura}. Given a single image, the goal is to synthesize a temporally coherent video that preserves visual identity while producing realistic motion. This task is inherently challenging since the model must infer plausible temporal dynamics from a single static frame, where motion, interactions, and physical evolution are fundamentally ambiguous. As a result, generating videos that are both temporally consistent and physically plausible remains difficult.

Recent work on controllable I2V generation has largely focused on trajectory-conditioned motion control~\cite{chuwan, wu2024draganything, shi2024motion, lei2025animateanything}. Methods such as Motion Prompting~\cite{geng2025motion}, Wan-Move~\cite{chuwan}, MagicMotion~\cite{li2025magicmotion}, and Motion-I2V~\cite{shi2024motion} guide diffusion-based video models using explicit motion signals, including point trajectories, optical flow, segmentation masks, or sparse bounding boxes. These signals are transformed into dense motion representations and injected into the generative backbone through mechanisms such as ControlNet~\cite{zhang2023adding} adapters, latent feature manipulation, or motion-aware attention, enabling user-specified object and camera motion while preserving appearance. Among these, Wan-Move achieves strong performance by directly modifying latent conditioning features of a large-scale Diffusion Transformer (DiT)~\cite{peebles2023scalable} model and training on millions of videos, improving motion fidelity and generalization. However, these methods treat motion primarily as geometric displacement, overlooking the physical causes governing object dynamics. To address this, a few physics-aware approaches introduce control signals based on forces or velocity fields to model object interactions \cite{gillmanforce,romero2025learning}, jointly model visual and latent physical dynamics~\cite{shen2026phantom}, or incorporate structured physical knowledge by injecting textual descriptions, qualitative categories, and quantitative physical properties through physics-aware attention~\cite{wangwisa}.\looseness-1

Despite recent progress, existing approaches for controllable video generation remain limited in how they model scene dynamics: \textbf{(1)} Current physics-aware approaches introduce control signals based on forces, velocities, or structured physical annotations \cite{romero2025learning, gillmanforce, wangwisa}. However, these methods typically encode physics as low-level control fields or predefined categories, which limits their ability to represent complex interactions between multiple entities in open-domain scenes. 
\textbf{(2)} Methods that condition on low-level geometric cues can make control cumbersome in multi-object scenarios, as users must author accurate signals for each object, and small errors in a few controls can yield globally implausible motion (\eg, inconsistent contact, sliding, penetrations, or desynchronized behavior). 
\textbf{(3)} Moreover, both trajectory-based and physics-conditioned methods rely heavily on large-scale curated datasets containing dense motion annotations, trajectories, optical flows, or physics labels, which are expensive to obtain and difficult to scale across diverse environments. For instance, Wan-Move~\cite{chuwan} is trained on 2M samples, Motion Prompting~\cite{geng2025motion} uses 2.2M, and Motion-I2V~\cite{shi2024motion} is trained on 10M samples. This dependence restricts the practicality and generalization of existing controllable video generation pipelines. 

To address these limitations and enable flexible interactive multi-object control, we introduce \modelnamenc, a graph-conditioned image-to-video generation framework that represents motion control through directed interaction scene graphs. Starting from an input image, GraphVid constructs a scene graph over detected entities with relational edges describing interactions (Figure \ref{fig:abstract_diagram}). Users interact directly with the image to specify desired dynamics, and the edited interaction graph is then translated into conditioning tokens that guide a frozen video diffusion backbone to generate a coherent video sequence.
To model interaction-driven dynamics, we introduce Edge-Aware Graph Reasoning, which conditions message passing on directed edge attributes so that object representations explicitly reflect their relational roles before being mapped to diffusion conditioning. By injecting this representation into a pretrained video DiT model and using LoRA modules in attention layers, the model learns to animate the scene using its own generative priors while respecting the specified interaction structure, enabling a more intuitive and interactive mechanism for motion control.
To support training, we construct \textsc{GraphVid-Bench}, an interaction-centric dataset of $\sim$27K video clips paired with explicit interaction graphs.
Despite using {$\sim$24$\times$} fewer parameters and substantially less training data, GraphVid achieves competitive performance with state-of-the-art motion-control video generation methods, significantly improving perceptual quality and reconstruction fidelity, and reducing FID by up to 39.9\% and FVD by 37.6\% compared to Motion-I2V, while improving PSNR 
(9.87→15.98) and SSIM (0.38→0.61). 
Our contributions are:
\begin{itemize}[itemsep=0.5ex, parsep=0pt, topsep=0pt, leftmargin=0.4cm]
\renewcommand{\labelitemi}{$\bullet$}
    \item We propose \modelnamenc, to the best of our knowledge, the first framework that enables interactive control of multi-object dynamics through directed interaction scene graphs constructed from a single image. GraphVid's graph-to-token conditioning interface injects interaction-aware representations into a frozen video diffusion transformer using parameter-efficient LoRA adaptation, enabling structured control without degrading pretrained generative priors.
   \item We introduce an \textbf{Edge-Aware Graph Reasoning module} that conditions graph message passing on directed edge attributes, allowing model representations to capture complex multi-object relations and interaction-driven motion propagation across multiple entities.
    \item  We curate \textbf{\textsc{GraphVid-Bench}}, a high-quality interaction-centric video generation dataset of $\sim$27K video clips paired with structured interaction graphs to support training and evaluation of relational video control.
    \item Despite requiring only 0.6B trainable parameters and achieving the fastest inference time among competing methods, GraphVid delivers substantial quality gains across perceptual and reconstruction metrics, 
    reducing FID by $34.5\%$ ($25.98\!\rightarrow\!17.02$) and FVD by $7.8\%$ ($107.89\!\rightarrow\!99.42$), while increasing PSNR by $5.9\%$ ($15.08\!\rightarrow\!15.98$) and SSIM by $15.0\%$ ($0.53\!\rightarrow\!0.61$) compared to WISA (physics-centric text inputs), and outperforming trajectory-based baselines such as Tora (FID $-30.3\%$, FVD $-10.0\%$, PSNR $+41.7\%$, SSIM $+12.9\%$) and
Motion-I2V (FID $-39.9\%$, FVD $-37.6\%$, PSNR $+61.9\%$, SSIM $+60.5\%$).
\end{itemize}

%% file: sections/2_related_work.tex
\section{Related Work}

\noindent \textbf{Video Generation.}
Recent advances in generative modeling have significantly improved the quality and scale of video generation systems~\cite{bar2024lumiere, yangcogvideox, wan2025wan}. Early approaches extended image diffusion models to the temporal domain using U-Net backbones and temporal modules to capture motion dynamics~\cite{guo2023sparsectrl, guo2023animatediff}. Subsequent work introduced latent video diffusion and motion-conditioned architectures to improve temporal consistency and scalability~\cite{hu2025lamd, yang2024motion}. 
More recently, large-scale diffusion transformers (DiTs) have emerged as the dominant architecture for high-quality video synthesis. Models such as CogVideoX~\cite{yangcogvideox} and Wan~\cite{wan2025wan} demonstrate that scaling transformer-based diffusion models enables long-range temporal coherence and improved visual fidelity. 
Recent works start from such strong pretrained video backbones and add a small number of trainable parameters that inject the desired control signal while preserving the backbone's generative priors~\cite{shi2024motion, chuwan, wei2025dreamrelation, li2026flashmotion}. This trend reflects both compute constraints and the empirical observation that large-scale backbones already encode rich appearance and temporal dynamics~\cite{geng2025motion, gillmanforce, lei2025animateanything, shen2026egoforge, susladkar2026pyratok}.
GraphVid follows this paradigm by keeping the video backbone frozen and learning lightweight modules that map structured interaction cues into conditioning tokens for generation.
\vspace{0.1cm}

\noindent \textbf{Controllable Video Generation.}
While general video generation models can synthesize realistic videos, controlling the generated content remains a major challenge. Early controllable generation approaches introduced conditioning signals such as pose, depth, or motion trajectories to guide video synthesis~\cite{shi2024motion, geng2025motion, lei2025animateanything}. Other works explored editing-based pipelines that modify existing videos while preserving temporal consistency~\cite{geyer2023tokenflow, ku2024anyv2v, li2024vidtome}. 
More recent methods focus on integrating structured conditioning signals or intermediate representations to enable more precise control. For example, models can condition on spatial layouts~\cite{he2024llms, wang2023videocomposer}, motion fields~\cite{lei2025animateanything, shi2024motion}, or object trajectories~\cite{wu2024draganything, yin2023dragnuwa} to guide generation while maintaining coherence with pretrained video backbones. 
Despite their strong performance, trajectory-based methods rely heavily on dense point-tracking supervision~\cite{karaev2024cotracker}, making them data-intensive and difficult to scale to higher-level relational reasoning and complex interactions between objects. In contrast, we propose GraphVid, which represents interactions as a directed scene graph and explicitly models multi-object control using an edge-aware GNN to generate more plausible and physically consistent videos. 
\vspace{0.1cm}

\noindent \textbf{Scene Graphs for Visual Generation.}
Scene graphs provide a structured representation of objects and their relations within a scene and have been widely studied for image understanding and generation~\cite{chang2104scene, li2024scene, dutta2025open, karwande2022chexrelnet}, 3D scene understanding~\cite{armeni20193d, wald2020learning, ogunleye20263d}, compositional and commonsense reasoning~\cite{khan2025survey, holla2024commonsense}, \etc, enabling models to reason about compositional structure and object interactions. 
Recent graph-to-video formulations support compositional generation from object-centric interaction graphs~\cite{bar2021compositional}.
In contrast, GraphVid targets \emph{interactive} control with an \emph{interaction-grounded} graph and learns lightweight modules that translate graph cues into conditioning tokens compatible with a video generation backbone, enabling structured control while retaining the backbone's strong priors.\looseness-1

%% file: sections/3_methodology.tex
\section{Method}
\FigMethod
We present \modelnamenc, a framework for controllable image-to-video (I2V) generation from a single conditioning image. 
The key idea is to represent user intent as a \emph{structured interaction graph} over entities in the scene, and to interface this graph with a large pretrained video diffusion transformer using a lightweight graph-to-token adapter and LoRA-based conditioning. 
By keeping the backbone frozen and learning only small modules, GraphVid introduces structured, object-centric control while preserving the strong generative priors of the pretrained model. 
Figure~\ref{fig:method_arch_diagram} presents an overview.

Given a conditioning image $I_0$, our goal is to generate a video sequence $\{I_t\}_{t=1}^{T}$ that follows a user-specified set of object interactions.
To express scene dynamics compositionally, we construct an attributed directed interaction graph $G=(V,E)$, where each node $v_i\in V$ corresponds to an entity detected in $I_0$ and each directed edge $e_{ij}\in E$ encodes a relation type $r_{ij}\in\mathcal{R}$ from object $i$ to object $j$ (\eg, \textit{push}, \textit{pull}, \textit{lift}, \textit{hold}),  describing how entities influence each other, and optional qualitative physical cues (\eg coarse direction and magnitude), allowing users to describe scene dynamics in intuitive relational terms. Unary directives that involve a single entity are represented as self-loop edges $e_{ii}$ (\eg, \texttt{rotate}, \texttt{move}, \texttt{scale}). 
We aim to learn a conditional video generator that models
$p(\mathbf{I}_{1:T}\mid I_0, G)$
where $\mathbf{I}_{1:T}=\{I_t\}_{t=1}^{T}$ denotes the $T$ generated frames.

\vspace{0.1cm}
\noindent \textbf{Node Representations.}
We extract object entities from the input frame $I_0$ using a pre-trained vision-language detector. For each detected object $v_i$, we construct a unified feature vector by fusing: 
(i) a visual embedding $\mathbf{f}^{vis}_i \in \mathbb{R}^{d_v}$ extracted by encoding the region of $I_0$ inside the object's bounding box $\mathbf{b}_i$ and pooling the resulting features, and
(ii) a text embedding $\mathbf{f}^{txt}_i \in \mathbb{R}^{d_t}$ encoding the object label. 
Normalized bounding box coordinates $\mathbf{b}_i \in \mathbb{R}^4$ are concatenated with these embeddings to form a representation $\mathbf{x}_i = [\mathbf{f}^{vis}_i \,\|\, \mathbf{f}^{txt}_i \,\|\, \mathbf{b}_i]$ that jointly captures object appearance, semantic identity, and spatial grounding.\looseness-1
\vspace{0.1cm}

\noindent \textbf{Edge Representations.}
Each directed edge $e_{ij}$ is represented with an open-vocabulary textual descriptor of the intended interaction between the source and target entities. 
Edges are encoded using a text embedding model to obtain the edge feature $\mathbf{e}_{ij} \in \mathbb{R}^{d_e}$.

\subsection{Edge-Aware Graph Reasoning}
\FigMethodLtx

Standard GCNs treat edges as binary connectivity signals and primarily propagate node features across the graph~\cite{kipf2016semi, hamilton2017inductive}. However, in physical interaction modeling, the type of relationship between objects directly determines their motion dynamics and implies fundamentally different motion patterns and causal effects. Ignoring these relational attributes reduces the graph to simple spatial adjacency and prevents the model from reasoning about how interactions influence object behavior. To address this, we employ an edge-aware Graph Isomorphism Network (GINEConv~\cite{xu2018powerful}) that explicitly incorporates edge semantics during message passing. As illustrated in Figure~\ref{fig:method_arch_ltx_diagram}, node and edge features are first projected into a shared hidden space $\mathbf{h}_i^{(0)}\!=\!\phi_n(\mathbf{x}_i), ~~
\mathbf{a}_{ij}\!=\!\phi_e(\mathbf{e}_{ij}),$
where $\phi_n$ and $\phi_e$ are learnable MLPs that map node/edge features to a common embedding space $\mathbb{R}^{d_h}$.
We then apply $L$ GINEConv layers to propagate relational information across the graph
\begin{equation}
\setlength{\abovedisplayskip}{7pt}
\setlength{\belowdisplayskip}{7pt}
\mathbf{h}_i^{(l+1)} =
\text{MLP}^{(l)}\left(
\mathbf{h}_i^{(l)} +
\sum_{j \in \mathcal{N}(i)}
\text{ReLU}(\mathbf{h}_j^{(l)} + \mathbf{a}_{ji})
\right).
\end{equation}
Unlike standard GCNs, the GINE formulation injects edge attributes directly into the aggregation step, allowing interaction semantics to influence node updates. Consequently, each node representation is conditioned not only on neighboring objects but also on the physical relationships governing their interactions. After $L$ layers, we obtain interaction-aware node embeddings $\mathbf{h}_i \in \mathbb{R}^{d_h}$ that encode both object identity and its dynamic relational state within the scene. These embeddings provide the downstream generative model with a structured representation of which entities should move and how they interact, enabling more physically coherent video generation.

\subsection{Graph-to-Video Adapter}
Modern Diffusion Transformers (DiTs) trained on large-scale video datasets provide strong spatio-temporal priors for video generation~\cite{yangcogvideox, lin2024open}. Training such models from scratch is computationally prohibitive and may compromise the rich generative capabilities learned from large-scale data. Therefore, we leverage a pretrained DiT backbone~\cite{hacohen2024ltx} and keep it frozen, introducing a lightweight adapter that converts graph embeddings into conditioning tokens compatible with the transformer's input space.

Each interaction-aware node embedding $\mathbf{h}_i$ is therefore projected into the transformer latent space as $\mathbf{z}_i\!=\text{MLP}(\mathbf{h}_i) \in \mathbb{R}^{d_l}$, where $d_l$ denotes the hidden size of the pretrained DiT. Each $\mathbf{z}_i$ is treated as a semantic conditioning token. 
Since scenes contain varying numbers of objects, node sequences are padded to a maximum length $N_{\max}$.
The resulting tokens, concatenated with edge text tokens, are passed as encoder hidden states. This enables the transformer’s self-attention to reason over entity interactions and translate graph structure into motion during generation. 

\subsection{Training Objective}

Since the pretrained DiT backbone is kept frozen to preserve its strong generative priors, the model requires a lightweight mechanism to adapt to the new graph conditioning modality. We therefore inject Low-Rank Adaptation (LoRA) \cite{hu2022lora} modules into the $\mathbf{Q}$, $\mathbf{K}$, $\mathbf{V}$, and output projection layers of each transformer block. Modulating these attention projections allows the model to incorporate graph tokens into the attention computation while introducing only a small number of trainable parameters. This enables the transformer to attend to interaction-aware graph embeddings without modifying the pretrained weights.\looseness-1

We train GraphVid using the flow-matching objective employed by the pretrained diffusion model. Given a target video $\mathbf{x}_0$, we encode it into the latent space and sample a timestep $t \sim p(t)$ from a logit-normal distribution. Noise is then added to obtain the noisy latent $\mathbf{x}_t$.
The model is optimized using loss
\begin{equation}
\small
\begin{aligned}
\mathcal{L}_{\mathrm{CFM}}(\theta)
&=
\mathbb{E}_{\mathbf{c}\sim \mathcal{I}_0, G}\,
\mathbb{E}_{t\sim p(t)}\,
\mathbb{E}_{\mathbf{x}_0\sim p_{\mathrm{data}}(\cdot\mid\mathbf{c}),\ \mathbf{x}_1\sim\mathcal{N}(0,\mathbf{I})}
\Big[
\big\|\mathbf{v}_\theta(\mathbf{x}_t,t,{\mathbf{c}})-(\mathbf{x}_1-\mathbf{x}_0)\big\|_2^2
\Big],
\end{aligned}
\end{equation}
where $\mathbf{x}_t=(1-t)\mathbf{x}_0+t\mathbf{x}_1$. During training, only the GNN parameters, Graph-to-Adapter MLP layers, and LoRA modules are updated (as $\theta$), while the VAE and DiT backbones remain frozen. This design enables end-to-end learning of a differentiable interactive/symbolic-to-latent mapping from interaction graphs to video dynamics while retaining the pretrained model’s generative capabilities.

%% file: sections/4_dataset.tex
\section{\textsc{GraphVid-Bench} Dataset}

Current video generation benchmarks primarily rely on text prompts~\cite{huang2024vbench} or low-level motion cues~\cite{chuwan}, which provide limited control over object-level interactions. 
However, many real-world dynamics are fundamentally relational, where actions emerge from how entities interact with one another (\eg, a person turning their head, an object moving relative to its surroundings). 
To enable structured control over such dynamics, we introduce \textsc{GraphVid-Bench}, a dataset that pairs videos with interaction graphs describing entities and their directed interaction cues.

\begin{wraptable}{r}{0.58\textwidth}
\vspace{-1cm}
\centering
\caption{\textbf{\textsc{GraphVid-Bench} statistics} across $\sim$27K curated videos. Interaction complexity highlights the prevalence of multi-entity dynamics.}
\small
\setlength{\tabcolsep}{5pt}
\resizebox{\linewidth}{!}{
\begin{tabular}{lccc}
\toprule
\textbf{Metric} & \textbf{Mean} & \textbf{Med.} & \textbf{Std.} \\
\midrule
Nodes / Graph & 7.76 & 5.0 & 8.03 \\
Edges / Graph & 4.85 & 4.0 & 3.46 \\
\midrule
\multicolumn{4}{c}{\textbf{Interaction Complexity}} \\
\midrule
\multicolumn{3}{l}{No-Interaction ($N{=}0$)} & 3881 \\
\multicolumn{3}{l}{Single-Interaction ($N{=}1$)} & 10982 \\
\multicolumn{3}{l}{Multi-Interaction ($N{>}1$)} & 12641 \\
\bottomrule
\end{tabular}
}
\vspace{-0.8cm}
\label{tab:dataset_stats}
\end{wraptable}\textsc{GraphVid-Bench} contains approximately \textbf{27K} interaction-centric video clips.  To ensure consistent training and evaluation, all clips are standardized to {81 frames} at {16 fps}. 
Each video is paired with a directed interaction graph where 
nodes represent scene entities and edges encode interaction cues between them. Node attributes include visual features, semantic labels, and spatial information, while edges capture the type and direction of interactions that influence scene dynamics.
As shown in Table~\ref{tab:dataset_stats}, the dataset exhibits diverse structural complexity, with an average of 7.76 nodes and 4.85 edges per graph. In total, 10,982 samples contain a single interaction, and 12,641 contain multiple interactions ($N>1$). 
As illustrated in Figure~\ref{fig:dataset_statistics}, \textsc{GraphVid-Bench} covers a broad spectrum of canonical physical interactions. The co-occurrence statistics further reveal that many interactions appear jointly within the same clips, indicating that real-world dynamics frequently involve multiple interacting primitives.
We also include 3,881 samples with no explicit interactions ($N\!=\!0$). These scenes contain natural object motion without direct entity-to-entity interactions, providing a baseline for learning general video dynamics. Since \modelnamenc allows users to add interaction edges at inference time, such samples help the model learn how scene behavior changes when new interactions are introduced. Appendix~\ref{app:dataset_details} provides additional details on the dataset curation process.
\FigDatasetStats

%% file: sections/5_experiments.tex
\section{Experiments}

\subsection{Implementation Details }
GraphVid leverages the pretrained LTX-Video~\cite{hacohen2024ltx} Diffusion Transformer as backbone for video generation. The  DiT backbone and the VAE encoder–decoder are kept frozen during training. Object entities are extracted from the conditioning image using Qwen3-VL~\cite{bai2025qwen3}. For each object, we obtain a $d_v = 8192$-dimensional visual embedding using mean and max pooling over object crops, a $d_t = 2560$-dimensional text embedding for the object label, and normalized bounding box coordinates. These are concatenated to form a $d_n = 10756$-dimensional node feature vector. Physical interactions between objects are encoded using Qwen3-Embedding~\cite{zhang2025qwen3}, producing edge attributes of dimension $d_e = 2560$. During edge-aware graph reasoning, node and edge features are projected to a hidden dimension of $d=512$, producing interaction-aware node embeddings. Node embeddings are projected to the LTX transformer hidden size using an MLP, mapping $512$-dimensional graph embeddings to $4096$ dimensions. Target videos are encoded into latent space using the pretrained VAE, and noise is injected at timesteps sampled from a logit-normal distribution. We employ LoRA modules of rank 128. During training, only the GNN parameters, graph-to-adapter projection layers, and LoRA weights are updated.

\begin{table}[t!]
\centering
\setlength{\tabcolsep}{5pt}
\caption{\textbf{Quantitative comparison on interaction-centric \textsc{MoveBench} subset.} GraphVid achieves competitive performance across all metrics while using significantly fewer parameters and training data. \colorbox{best}{Best}  and \colorbox{second}{second-best} are highlighted.}
\label{tab:distilled-movebench}
\vspace{-4mm}
\renewcommand{\arraystretch}{1.1}
\resizebox{\linewidth}{!}{
\begin{tabular}{l ccc ccccc}
\toprule
\textbf{Method} & \shortstack{\textbf{Train}\\\textbf{Data}} & \shortstack{\textbf{Trainable}\\\textbf{Params (B)}} & \shortstack{\textbf{Inference}\\\textbf{(s)}$\downarrow$} & \textbf{FID}$\downarrow$ & \textbf{FVD}$\downarrow$ & \textbf{PSNR}$\uparrow$ & \textbf{SSIM}$\uparrow$  & \textbf{EPE}$\downarrow$ \\
\midrule

\rowcolor{grayrow}
Wan-Move~\cite{chuwan} & 2M & 14.5 & 1800 & 15.56 & 82.17 & 17.21 & 0.61 & 2.6 \\
Motion-I2V~\cite{shi2024motion} & 10M & 1.2 & 790 & 28.32 & 159.32 & 9.87 & 0.38 & 3.9 \\

Tora~\cite{zhang2025Tora} & 630K & 5 & 1200 & 24.45 & 110.47 & 11.27 & 0.54 & 3.3\\

MagicMotion~\cite{li2025magicmotion} & 23K & 1.5 & 750 & 26.57 & 105.12 & 12.04 & 0.51 & \cellcolor{second}3.2\\

WISA~\cite{wangwisa} & 80K & 1 & 1000 & 25.98 & 107.89 & \cellcolor{second}15.08 & 0.53 & 4.1\\

FlashMotion~\cite{li2026flashmotion} & 23K & 13 & \cellcolor{second}677 & \cellcolor{second}19.02 & \cellcolor{second}104.12 & 14.04 & \cellcolor{second}0.56 & 3.8\\

\modelnamenc (\textbf{Ours}) & 27K & 0.6 & \cellcolor{best}200 & \cellcolor{best}17.02 & \cellcolor{best}99.42 & \cellcolor{best}15.98 & \cellcolor{best}0.61 & \cellcolor{best}2.9\\

\bottomrule
\end{tabular}
}
\vspace{-0.1cm}
\end{table}

\subsection{Experimental Setup}
For evaluation, we establish a \textsc{MoveBench}~\cite{chuwan} subset focused on interaction-centric scenarios as the primary evaluation suite. Since GraphVid emphasizes physics-aware relational control rather than explicit pixel-level motion supervision, we prioritize metrics that assess overall video quality, temporal coherence, and structural fidelity. In addition, we evaluate performance on a subset of the distilled \textsc{MoveBench} containing only multi-object interaction scenarios, allowing us to assess the effectiveness of GraphVid in more complex settings where reasoning over interactions between multiple entities is required. We report Frechet Video Distance (FVD) \cite{unterthiner2019fvd}, Frechet Inception Distance (FID) \cite{heusel2017gans}, Peak Signal-to-Noise Ratio (PSNR), and Structural Similarity Index (SSIM) \cite{wang2004image} to measure visual quality.  We additionally report End-Point Error (EPE)~\cite{geng2025motion}, the average displacement between predicted and ground-truth optical flow, which directly measures whether the generated motion follows the intended interactions. Additional details can be found in Appendix~\ref{app:addn_exp_detials}.

\subsection{Quantitative Results}
Table~\ref{tab:distilled-movebench} compares GraphVid with existing controllable video generation methods on the interaction-centric \textsc{MoveBench} benchmark subset, where GraphVid achieves competitive performance across all metrics while requiring substantially fewer training samples, significantly fewer trainable parameters, and faster inference.
Among prior approaches, WISA~\cite{wangwisa} is the most closely related method that explicitly incorporates structured physical knowledge into the video generation process. Compared to WISA, GraphVid achieves a substantial improvement in perceptual quality and reconstruction fidelity. In particular, GraphVid reduces FID from 25.98 to 17.02 and FVD from 107.89 to 99.42, corresponding to a 34.5\% and 7.8\% improvement, respectively, indicating significantly better visual realism and distribution alignment with ground-truth videos. Similarly, SSIM improves by 15.0\%, and PSNR by 5.9\%, demonstrating that our graph-based interaction modeling produces frames that are both sharper and structurally more consistent. 
This advantage also holds for motion accuracy, with GraphVid reducing EPE from 4.1 to 2.9, a 29.3\% improvement over WISA, and achieving the lowest EPE among all
comparable-scale methods. Since EPE measures how closely the generated optical flow matches the ground truth, this indicates that our graph conditioning produces motion that most faithfully follows the specified interactions. The overall perceptual, reconstruction, and motion gains indicate that explicitly modeling interaction relationships through a directed scene graph provides a stronger inductive bias for controllable video generation than relying on predefined textual ({physics-centric}) attributes.

\begin{table}[t!]
\centering
\caption{\textbf{Multi-object quantitative comparison on interaction-centric \textsc{MoveBench}.} GraphVid maintains strong structural coherence compared to larger baselines.  \colorbox{best}{Best}  and \colorbox{second}{second-best} are highlighted.}
\label{tab:distilled-movebench-multi}
\vspace{-4mm}
\setlength{\tabcolsep}{10pt}
\renewcommand{\arraystretch}{1.1}
\resizebox{0.99\linewidth}{!}{
\begin{tabular}{l cc ccccc}
\toprule
\textbf{Method} & \shortstack{\textbf{Train}\\\textbf{Data}} & \shortstack{\textbf{Trainable}\\\textbf{Params (B)}} & \textbf{FID}$\downarrow$ & \textbf{FVD}$\downarrow$ & \textbf{PSNR}$\uparrow$ & \textbf{SSIM}$\uparrow$  & \textbf{EPE}$\downarrow$ \\
\midrule
\rowcolor{grayrow}
Wan-Move~\cite{chuwan} & 2M & 14.5 & 31.29 & 252 & 16.69 & 0.61 & 2.2 \\

Tora~\cite{zhang2025Tora} & 630K & 5 & 56.04 & 369 & \cellcolor{best}14.98 & \cellcolor{second}0.52 &  \cellcolor{second}3.5 \\

WISA~\cite{wangwisa} & 80K & 1 & 60.12 & 341 & 13.13 & \cellcolor{best}0.55 & 3.9 \\

FlashMotion~\cite{li2026flashmotion} & 23K & 13 & \cellcolor{second}55.03 & \cellcolor{second}311 & 12.12 & \cellcolor{second}0.52 & 3.9 \\

\modelnamenc (\textbf{Ours})  & 27K & 0.6 & \cellcolor{best}49.45 & \cellcolor{best}291 & \cellcolor{second}14.44 & \cellcolor{best}0.55 &  \cellcolor{best}3.0 \\

\bottomrule
\end{tabular}
}
\vspace{-0.1cm}
\end{table}
We further compare against trajectory-based control methods such as Motion-I2V~\cite{shi2024motion}, MagicMotion~\cite{li2025magicmotion}, FlashMotion~\cite{li2026flashmotion}, and Tora~\cite{zhang2025Tora}. These approaches rely heavily on explicit motion signals such as trajectories, optical flow, or segmentation masks. Compared to Motion-I2V, GraphVid reduces FID by 39.9\% (28.32 $\rightarrow$ 17.02) and FVD by 37.6\% (159.32 $\rightarrow$ 99.42), while improving PSNR from 9.87 to 15.98 and SSIM from 0.38 to 0.61, indicating substantially better frame reconstruction and perceptual fidelity. Compared to FlashMotion, a recent trajectory-guided video generation method, GraphVid lowers FID by 10.5\% (19.02 $\rightarrow$ 17.02) while also achieving stronger reconstruction quality, improving PSNR from 14.04 to 15.98 and SSIM from 0.56 to 0.61.
Compared with MagicMotion, GraphVid reduces FID by approximately 36\% (26.57 → 17.02) and improves temporal consistency. 
Notably, GraphVid attains a lower EPE (2.9) than all comparable trajectory-based methods
(Motion-I2V 3.9, FlashMotion 3.8, Tora 3.3, MagicMotion 3.2), reducing motion
error by 9.4--25.6\% without relying on any explicit motion supervision.
These improvements highlight the effectiveness of modeling scene dynamics through structured interaction reasoning. Wan-Move~\cite{chuwan} achieves the best absolute metrics due to its significantly larger scale, being trained on approximately 2M videos and using a 14.5B parameter backbone. In contrast, GraphVid operates with only 0.6B trainable parameters and is trained on a much smaller 27K interaction-centric dataset. Despite using approximately two orders of magnitude less training data and $\sim$24$\times$ lower model params, GraphVid achieves competitive performance, outperforming most prior controllable methods on reconstruction and
motion-accuracy.

A similar trend can be observed in Table~\ref{tab:distilled-movebench-multi}, which evaluates performance on the multi-object subset of \textsc{MoveBench} where interaction reasoning becomes particularly important. 
In these scenarios, GraphVid improves multi-object interaction quality over other lightweight baselines, with clear relative gains, while using substantially fewer resources. When compared to models of similar data scale, GraphVid achieves the best FID (49.45), improving over the next best (FlashMotion, 55.03) by 10.1\%, and the best FVD (291), improving over the next best (FlashMotion, 311) by 6.4\%; both indicating better perceptual realism and temporal consistency in multi-entity dynamics. 
GraphVid also achieves the lowest EPE among comparable-scale methods, confirming that its motion-accuracy advantage carries over to the harder multi-object setting. 
On reconstruction fidelity, GraphVid remains competitive with Tora (best PSNR) and WISA (best SSIM). Importantly, these gains come with far lower scale. GraphVid uses 0.6B trainable parameters ($\approx$21.7$\times$ fewer than FlashMotion and 8.3$\times$ fewer than Tora) and trains on 27K clips (comparable to FlashMotion and 23$\times$ fewer than Tora), underscoring that interaction-graph conditioning provides strong multi-object coherence without requiring massive models or data. Appendix~\ref{app:davis_experiments} presents additional results on our interaction-focused DAVIS~\cite{pont20172017} subset. \looseness-1

\begin{wrapfigure}{r}{0.5\linewidth}
    \vspace{-7mm}
    \centering
    \includegraphics[width=\linewidth]{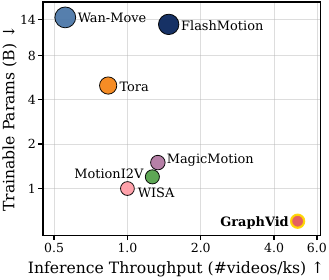}
    \vspace{-7mm}
    \caption{\textbf{Model efficiency comparison.} Trade-off between inference throughput and scale. Bubble size reflects model scale.}
    \label{fig:efficiency_tradeoff_movebench}
    \vspace{-6mm}
\end{wrapfigure}\subsection{Efficiency and Scalability Analysis} 
\Cref{fig:efficiency_tradeoff_movebench} plots inference throughput against model scale. Beyond accuracy, GraphVid offers strong efficiency advantages, achieving
the fastest inference among all compared approaches at 200 seconds, significantly
faster than Wan-Move (1800s), Tora (1200s), and WISA (1000s).
This efficiency stems from our lightweight interaction graph
conditioning and LoRA-based integration into the pretrained DiT backbone, 
which avoids the need for heavy auxiliary motion encoders or large-scale motion prediction modules.
This confirms that explicitly modeling interaction relationships through a structured scene graph provides a powerful and efficient alternative to trajectory-based motion supervision. By leveraging interaction-aware graph reasoning, GraphVid achieves strong controllability and physical plausibility while operating with substantially lower compute and data.

\FigQualitative
\subsection{Qualitative Analysis}
Figure~\ref{fig:graphvid_qualitative_results} presents representative qualitative results demonstrating that modeling scene dynamics through GraphVid interaction graphs enables precise and interpretable control over complex multi-entity motion. Examples illustrate diverse interaction types, including object–object interactions, human–object interactions, and articulated human motion. Across all scenarios, GraphVid preserves object identity and scene appearance, produces realistic motion trajectories and interaction-consistent behavior, and maintains stable background geometry. 
These examples indicate that the interaction-graph interface helps the model bind motion to the correct entities and combine multiple primitives into coherent multi-step dynamics, something that is typically brittle with purely text- or trajectory-only controls. Appendix~\ref{app:human_pref_study} presents a human preference study, while Appendix~\ref{app:addn_qual_results} provides qualitative results across diverse interaction scenarios.\looseness-1

\subsection{Ablation Studies}

\noindent \textbf{Backbone Generalization.}
To verify that relational scene graphs serve as a 
universal conditioning modality rather
than an exploit of LTX's specific latent 
\begin{wraptable}{r}{0.6\textwidth}
\vspace{-1.2cm}
\centering
\setlength{\tabcolsep}{4pt}
\caption{\textbf{Backbone ablation.}}
\label{tab:ablation-backbone}
\vspace{-1mm}
\resizebox{\linewidth}{!}{
\begin{tabular}{lccccc}
\toprule
\textbf{Backbone} & \textbf{FID}$\downarrow$ & \textbf{FVD}$\downarrow$ & \textbf{PSNR}$\uparrow$ & \textbf{SSIM}$\uparrow$  & \textbf{EPE}$\downarrow$ \\
\midrule
Ours + LTX & \textbf{17.02} & \textbf{99.42} & \textbf{15.98} & \textbf{0.61} & \textbf{2.9}  \\
Ours + Wan 2.2 & 17.29 & 100.01 & 15.70 & 0.60 & 3.1 \\
\bottomrule
\end{tabular}}
\vspace{-0.9cm}
\end{wraptable}priors, we replace the 2B LTX backbone with the 5B Wan 2.2 model\cite{wan2025wan}. When training with this variant, our edge-aware GNN and adapter successfully map the topological graphs into Wan's latent space, achieving competitive performance (\Cref{tab:ablation-backbone}). This demonstrates that our graph representation is a backbone-agnostic conditioning signal capable of guiding scene dynamics across different foundation models.
\vspace{0.1cm}

\begin{wraptable}{r}{0.5\textwidth} 
\vspace{-1cm} 
\centering 
\setlength{\tabcolsep}{8pt} 
\caption{\textbf{Graph ablation.}} \label{tab:ablation-graph-density} \vspace{-1mm} 
\resizebox{\linewidth}{!}
{ \begin{tabular}{ccccc} 
\toprule \#\textbf{Nodes} & \textbf{FID}$\downarrow$ & \textbf{FVD}$\downarrow$ & \textbf{PSNR}$\uparrow$ & \textbf{SSIM}$\uparrow$\\ 
\midrule 10 & 17.25  & 102.04   & 15.56  & 0.60 \\
30 & \textbf{17.02} & \textbf{99.42} & \textbf{15.98} & \textbf{0.61}\\ 
50 & 17.17 & 104.97 & 15.11 & 0.60\\ 
128 & 17.55 & 105.45 & 14.55 & 0.59 \\ 
256 & 17.97 & 109.92 & 13.77 & 0.57 \\ 
\bottomrule 
\end{tabular}} 
\vspace{-0.8cm} 
\end{wraptable}\noindent\textbf{Graph Sparsity and Attention Dilution.}
We evaluate the effect of the context window size on the DiT's self-attention mechanism by varying the maximum node capacity. As shown in Table~\ref{tab:ablation-graph-density}, padding sparse real-world scene graphs to 
overly large context windows can dilute attention and degrade generation quality. 
Capping the graph size at 30 nodes consistently outperforms larger capacities (128/256), achieving the best FVD (99.42). This suggests that a spatially tighter, denser node set yields better structural alignment: as the maximum node budget increases (\eg, from 30 → 256), FID degrades, likely because the additional capacity is dominated by zero-padding rather than informative structure. The resulting padded tokens inject noise into attention and dilute meaningful relational cues, making it harder for the transformer to concentrate on active interactions. Overall, these results suggest that balanced, densely populated graph representations provide the strongest conditioning signal, and GraphVid therefore uses a maximum node capacity of 30 for all experiments.

\noindent \textbf{Conditioning Modalities and Edge Semantics.}
To evaluate the importance of structured conditioning and edge semantics, we perform an inference-time ablation comparing three distinct regimes. Table~\ref{tab:ablation-conditioning} shows (1) the standard LTX 
\begin{wraptable}{r}{0.65\textwidth}
\vspace{-1.1cm}
\centering
\setlength{\tabcolsep}{4pt}
\caption{\textbf{Conditioning modalities.}}
\label{tab:ablation-conditioning}
\vspace{-1mm}
\resizebox{\linewidth}{!}{
\begin{tabular}{lcccccc}
\toprule
\textbf{Variant} & \textbf{FID}$\downarrow$ & \textbf{FVD}$\downarrow$ & \textbf{PSNR}$\uparrow$ & \textbf{SSIM}$\uparrow$  & \textbf{EPE}$\downarrow$ \\
\midrule
Text + Img (LTX) & 20.04 & 109.34 & 11.23 & 0.53 & 3.6 \\
Graph (no edge text) & 17.52 & 101.32 & \textbf{16.88} & 0.60 & 3.0 \\
Graph (full) & \textbf{17.02} & \textbf{99.42} & 15.98 & \textbf{0.61} & \textbf{2.9} \\
\bottomrule
\end{tabular}}
\vspace{-0.8cm}
\end{wraptable}baseline, which generates video using a global text prompt and an initial image; (2) GraphVid conditioned on the image and a topology-only graph, where the user provides interaction arrows and overlays but omits textual edge context; and (3) our full GraphVid approach, utilizing the image and a complete graph equipped with explicit text context for each interaction edge. While the text-based baseline achieves moderate visual quality, it struggles to bind localized motion to specific entities (FID 20.04). Replacing the global text with our topology-only graph improves FID to 17.52, indicating that explicitly graphing spatial boundaries aids motion localization. However, without the semantic richness of the textual interaction embeddings, the model exhibits severe motion ambiguity, failing to distinguish opposing interactions like ``pushing'' and ``pulling''. The full graph conditioning outperforms both baselines, reducing FID to 17.02 and FVD to 99.42. These results suggest that graph topology alone provides useful spatial grounding, while semantic edge labels supply additional interaction intent that helps disambiguate object relationships and improve motion generation.

\begin{wraptable}{r}{0.6\linewidth}
\centering
\vspace{-1cm}
\caption{\textbf{Ablation on LoRA rank.}}
\label{tab:rank_ablation}
\vspace{-1mm}
\setlength{\tabcolsep}{8pt}
\resizebox{\linewidth}{!}{
\begin{tabular}{ccccc}
\toprule
\textbf{Rank} & \textbf{FID}$\downarrow$ & \textbf{FVD}$\downarrow$ & \textbf{PSNR}$\uparrow$ & \textbf{SSIM}$\uparrow$\\
\midrule
16  & 18.34 & 103.18 & 13.70 & 0.58 \\
32  & 17.89 & 102.39 & 14.51 & 0.60 \\
64  & 17.53 & 100.99 & 15.25 & 0.60 \\
128 & \textbf{17.02} & \textbf{99.42} & \textbf{15.98} & \textbf{0.61} \\
\bottomrule
\end{tabular}
}
\vspace{-0.8cm}
\end{wraptable}\noindent \textbf{LoRA Rank and Capacity Trade-offs.}
We inject low-rank adaptation (LoRA) matrices into the transformer blocks to modulate the pre-trained weights with our graph conditions. In this ablation, we study the effect of the LoRA rank on generation quality and reconstruction fidelity. As shown in Table~\ref{tab:rank_ablation}, increasing the rank consistently improves performance across all metrics. Moving from rank 16 to 128 reduces FID from 18.34 to 17.02 (–7.2\%) and FVD from 103.18 to 99.42 (–3.6\%), indicating better perceptual quality and temporal coherence. At the same time, reconstruction metrics improve substantially, with PSNR increasing from 13.70 to 15.98 (+16.6\%) and SSIM from 0.58 to 0.61 (+5.2\%). These results suggest that higher-rank adapters provide greater representational capacity to capture complex motion and appearance variations, leading to more faithful and temporally consistent video generation. Rank 128 is used in all the reported experiments.

%% file: sections/6_conculsion.tex
\section{Conclusion}

We introduce \modelnamenc, a framework for controllable image-to-video generation that represents scene dynamics through directed interaction graphs. Unlike prior approaches that rely on dense motion signals or low-level physics annotations, GraphVid models motion as structured interactions between entities, enabling intuitive multi-object control directly from a single image. By integrating Edge-Aware Graph Reasoning with a frozen video diffusion transformer through parameter-efficient LoRA conditioning, GraphVid preserves strong pretrained generative priors while enabling structured control over scene dynamics. To support this paradigm, we also curate \textbf{\textsc{GraphVid-Bench}}, a dataset of 27K interaction-centric videos paired with explicit interaction graphs, providing supervision for learning relational video dynamics. Experiments show that GraphVid matches or outperforms prior controllable video generation methods while using far fewer parameters and training data, highlighting interaction graphs as a scalable interface for controllable video generation.

\section*{Acknowledgments}
This research was partially supported by Google, the Google TPU Research Cloud (TRC) program, the NSF CAREER Award \#2542328, the U.S. Defense Advanced Research Projects Agency (DARPA) under award HR001125C0303, and the U.S. Army under contract W5170125CA160. The views and conclusions contained herein are those of the authors and should not be interpreted as necessarily representing the official policies, either expressed or implied, of Google, NSF, DARPA, the U.S. Army, or the U.S. Government. The U.S. Government is authorized to reproduce and distribute reprints for governmental purposes notwithstanding any copyright annotation therein. This work also used the Delta GPUs provided by the National Center for Supercomputing Applications through allocations CIS240808 and CIS250318 from the Advanced Cyberinfrastructure Coordination Ecosystem: Services \& Support (ACCESS \cite{boerner2023access}) program, supported by National Science Foundation grants \#2138259, \#2138286, \#2138307, \#2137603, and \#2138296.

%% file: sections/7_supplementary.tex

\clearpage
\appendix
\phantomsection
\section{\textsc{GraphVid-Bench} Details and Curation Pipeline}
\label{app:dataset_details}

\noindent \textbf{Dataset Provenance and Aggregation.}
We construct our \textsc{GraphVid-Bench} training data from three complementary open-source datasets to ensure a broad distribution of actions and interaction types. Specifically, we use WISA-80K~\cite{wangwisa} for rigid-body physics-driven dynamics, \textsc{Something-Something} v2~\cite{goyal2017something} for diverse commonsense human-object and object-object interactions, and MagicData~\cite{li2025magicmotion} for complex real-world foreground motion and multi-interaction scenarios.
For evaluation, we curate dedicated test splits distilled from \textsc{DAVIS} 2017~\cite{pont20172017} and \textsc{MoveBench}~\cite{chuwan} to assess generalization on established benchmarks.\looseness-1

\noindent \textbf{Automated Semantic Filtering.}
To remove confounding motion factors, we introduce an automated semantic filtering stage using a vision-language model (Qwen3-VL~\cite{bai2025qwen3}). For each video, lightweight proxy clips are extracted and provided to the vision-language model to classify the dominant motion source into categories such as camera motion, ego-motion, ambient environmental dynamics, or discrete object interactions.
We retain only videos classified as exhibiting discrete object interactions and discard sequences dominated by camera panning, background motion, or fluid/gaseous environmental effects. This filtering step ensures that the final dataset emphasizes interaction-driven scene dynamics rather than camera-induced motion.

\noindent \textbf{Spatio-Temporal Standardization and Motion-Energy Windowing.}
To align the dataset with the latent resolution constraints of the video diffusion backbone and to maintain consistent training inputs, all videos are standardized spatially and temporally. Each video is resized to a fixed resolution of $512 \times 288$. Instead of center-cropping, which often truncates key interactions near frame boundaries, we apply dynamic letterboxing to preserve the original aspect ratio and avoid geometric distortion of interacting entities.
The temporal dimension is standardized to 81 frames at 16 FPS. Since many source videos exceed this duration, naive random cropping may capture segments where the interaction is inactive. To address this, we introduce a sliding motion-energy window algorithm. 
The preprocessing pipeline computes a motion-energy signal by summing the pixel-wise absolute differences between consecutive grayscale frames. An 81-frame window is then slid across this signal, and the segment with the highest aggregated motion energy is selected. This ensures that the extracted clip corresponds to the most active phase of the interaction, providing stronger supervisory signals for training.
During training, a random subset of each batch is upscaled to different backbone-compatible resolutions, encouraging the model to generate high-quality videos across multiple output resolutions.

\subsection{Test Set Distillation}

\FigMoveBenchSubset

To ensure fair and consistent evaluation, the test datasets undergo the same spatio-temporal standardization pipeline described above.

\noindent \textbf{\textsc{DAVIS} 2017:} Raw frame sequences are first converted into video tensors. Sequences slightly shorter than 81 frames (\eg, 78–79 frames) are edge-padded using the final frame to satisfy the temporal requirement. Longer sequences are processed using the motion-energy windowing procedure to extract the most dynamic segment. The resulting clips are then filtered using the VLM motion classifier to remove sequences dominated by camera or ego-motion.

\noindent \textbf{\textsc{MoveBench}:} Videos are first filtered by the VLM to retain only clips exhibiting discrete object-level physical interactions. The retained videos are then resized to $512 \times 288$ and standardized to 81 frames to match the visual input requirements of the downstream model. Figure~\ref{fig:movebench_subset_distribution} shows the category distribution of both the original \textsc{MoveBench} dataset and our distilled subset. 

\section{Additional Experimental Details}
\label{app:addn_exp_detials}
We train our Edge-Aware GNN~\cite{xu2018powerful} with a hidden dimension of 512. The graph representation is projected into the 4096-dimensional latent space of the LTX 2B backbone~\cite{hacohen2024ltx} through a 3-layer MLP adapter, while the backbone itself remains frozen throughout training. To enable parameter-efficient adaptation, we insert LoRA modules into the attention projections, namely \texttt{to\_q}, \texttt{to\_k}, \texttt{to\_v}, and \texttt{to\_out.0}, with rank $r=128$ and scaling factor $\alpha=32$.
Training is performed on 8 NVIDIA A100 GPUs for approximately 2.5 days using \texttt{bfloat16} mixed precision and the Accelerate library for memory-efficient distributed optimization. We use AdamW with a learning rate of $1\times10^{-4}$. Unless otherwise stated, inference is conducted at a spatial resolution of $512\times288$, with 81 frames at 16 FPS. For sampling, we adopt the \texttt{FlowMatchEulerDiscreteScheduler} with continuous time-step sampling. For fair comparison with prior methods, FlashAttention is disabled during inference.

\noindent \textbf{Environmental Impact.}
We estimate the training carbon footprint using a uniform hardware-level accounting protocol for both GraphVid and Wan-Move. Assuming the maximum thermal design power of an NVIDIA A100 GPU (400\,W), our training setup with 8 A100 GPUs for approximately 2.5 days (60 hours) corresponds to a GPU-only energy consumption of $8 \times 0.4 \times 60 = 192~\text{kWh}.$
Under the same assumption, Wan-Move, which uses 64 A100 GPUs for 10 days (240 hours), corresponds to $64 \times 0.4 \times 240 = 6144~\text{kWh}.$

\section{Additional Experiments}
\label{app:davis_experiments}

In this section, we provide extended empirical evaluations to further validate GraphVid's efficiency, generalization capability, and architectural robustness.

\begin{table}[t!]
\centering
\setlength{\tabcolsep}{9pt}
\renewcommand{\arraystretch}{1.05}
\caption{\textbf{Quantitative comparison on the \textsc{DAVIS} motion benchmark.}
We evaluate controllable video generation quality across perceptual (FID, FVD)
and reconstruction (PSNR, SSIM) metrics. GraphVid achieves strong overall
performance while using significantly fewer trainable parameters.
Highlighted \colorbox{best}{best} and \colorbox{second}{second-best} among the parameter-efficient architectures.}
\label{tab:davis_results}
\vspace{-0.4cm}
\resizebox{0.99\linewidth}{!}{
\begin{tabular}{l c ccccc}
\toprule
\textbf{Method} &  \shortstack{\textbf{Trainable}\\\textbf{Params (B)}} & \textbf{FID}$\downarrow$ & \textbf{FVD}$\downarrow$ & \textbf{PSNR}$\uparrow$ & \textbf{SSIM}$\uparrow$  & \textbf{EPE}$\downarrow$ \\
\midrule
\rowcolor{grayrow}
Wan-Move~\cite{chuwan}
& 14.5
& 15.04
& 86
& 12.37
& 0.57 
& 2.5 \\

Tora~\cite{zhang2025Tora}
& 5.0
& 24.56
& 107
& 10.21
& \cellcolor{second}0.49
& 3.5 \\

Motion-I2V~\cite{shi2024motion}
& 1.2
& 26.12
& 114
& 9.98
& 0.43
& 3.8 \\

WISA~\cite{wangwisa}
& 1.0
& 25.05
& 110
& 10.45
& 0.47
& \cellcolor{second}3.3 \\

MagicMotion~\cite{li2025magicmotion}
& 1.5
& 20.11
& \cellcolor{second}100
& \cellcolor{second}11.32
& \cellcolor{second}0.49
& 3.5 \\

FlashMotion~\cite{li2026flashmotion}
& 13
& \cellcolor{second}17.09
& 107
& 9.92
& \cellcolor{second}0.49
& 3.4 \\

\modelnamenc \textbf{(Ours)}
& \textbf{0.6}
& \cellcolor{best}16.20
& \cellcolor{best}98
& \cellcolor{best}12.85
& \cellcolor{best}0.55
& \cellcolor{best}2.9 \\

\bottomrule
\end{tabular}
}
\vspace{-1mm}
\end{table}

\noindent \textbf{Evaluation on the filtered \textsc{DAVIS} benchmark.}
\Cref{tab:davis_results} evaluates  GraphVid on our interaction-centric \textsc{DAVIS} subset, where GraphVid achieves highly competitive performance while utilizing a fraction of the trainable parameters (0.6B) and training data (27K videos).
Compared to WISA~\cite{wangwisa}, GraphVid substantially improves both temporal coherence (FVD: 110 $\rightarrow$ 98) and perceptual quality (FID: 25.05 $\rightarrow$ 16.20). This confirms that explicitly modeling interaction relationships through a directed scene graph provides a much stronger inductive bias than relying on predefined textual attributes. Furthermore, GraphVid consistently achieves better structural fidelity (PSNR: 12.85, SSIM: 0.55), perceptual realism, and temporal consistency (FVD: 98) than trajectory-conditioned methods (\ie, Motion-I2V~\cite{shi2024motion}, Tora~\cite{zhang2025Tora}, MagicMotion~\cite{li2025magicmotion}, and FlashMotion~\cite{li2026flashmotion}). GraphVid excels in overall frame reconstruction, highlighting the advantage of structured interaction reasoning over geometric motion cues alone.
Finally, while the heavyweight Wan-Move (14.5B parameters, 2M training videos) establishes the best performance, GraphVid achieves state-of-the-art perceptual and structural fidelity among all parameter-efficient baselines despite an order-of-magnitude reduction in model capacity and training scale.

\begin{table}[t!]
\centering
\setlength{\tabcolsep}{6pt}
\renewcommand{\arraystretch}{1.08}
\caption{\textbf{Training-time graph representation ablation.} We compare graph variants that use only node features against our GraphVid full model, which additionally incorporates semantic edge embeddings through message passing.}
\label{tab:ablation-graph-repr}
\vspace{-0.35cm}
\begin{tabular}{lccccc}
\toprule
\textbf{Variant} & \textbf{Nodes} & \textbf{Edges} & \textbf{FID}$\downarrow$ & \textbf{FVD}$\downarrow$ & \textbf{PSNR}$\uparrow$ \\
\midrule
Node-only Graph Encoder & {\color{green}\checkmark} & {\color{red}\ding{55}} & 19.55 & 103.00 & 15.24 \\
\textbf{GraphVid (Ours)} & {\color{green}\checkmark} & {\color{green}\checkmark} & \textbf{17.02} & \textbf{99.42} & \textbf{15.98} \\
\bottomrule
\end{tabular}
\vspace{-2mm}
\end{table}
\noindent \textbf{Edge-Aware GNN Training Robustness.}
We conduct an architectural ablation to isolate the contribution of relational edge messaging within our GNN. 
We train a GraphVid baseline variant utilizing only isolated node features (extracted bounding boxes and visual object embeddings) and omitting the semantic edge connections and message-passing layers. We compare this against our full Edge-Aware GNN implementation, which inherently integrates the relational dynamics through its explicit edge-routing architecture. 
As detailed in Table~\ref{tab:ablation-graph-repr}, while the node-only variant converges and provides a baseline level of control (FVD 103), it fails to fully resolve the complex physical dependencies between interacting entities. The integration of explicit edge features during training substantially improves generation quality across all metrics, driving FVD down to 99.42 and improving structural reconstruction (PSNR 15.98). This confirms that learning the physical causality of a scene fundamentally requires modeling the relationships {between} objects.\looseness-1

\section{Human Preference Study}
\label{app:human_pref_study}
We conduct a human evaluation study to assess how well the generated videos satisfy the intended interaction semantics described in the prompt. The study includes a diverse group of 60 participants evaluating 10 distinct interaction scenarios. Participants are presented with the text prompt describing the intended interaction, the first-frame conditioning image, and a short motion guidance video illustrating the desired trajectory. They are then shown four anonymized candidate videos and asked two separate questions to select the result that (a) best satisfies the semantic intent of the prompt and (b) has the best visual quality. For each scenario, participants provide preference judgments, yielding $60 \times 10 \times 2 = 1200$ individual judgments. 

In Figure~\ref{fig:overall_comparison}, we observe that \modelname{} achieves a high Win Rate of 60\% and 90\% for semantic intent and visual quality, respectively. 
Figure~\ref{fig:overall_comparison} also summarizes the aggregated preference rates for models that were preferred for both metrics. Across the 10 diverse evaluated scenarios, \modelnamenc wins the majority vote for \textit{both} semantic intent and visual quality in {6 out of 10} test cases, with 3 ties and only 1 loss. Moreover, \modelname{} receives the highest semantic preference, capturing {39.5\%} of all votes. This substantially outperforms Wan-Move (22.3\%), Tora (19.3\%), and WISA (18.9\%), representing a relative improvement of over 77\% compared to the strongest baseline.
The strong preference for GraphVid confirms that users perceive the generated videos as better capturing the core intent of the prompt, bridging the gap between abstract control signals and visually realistic video generation. These consistent results indicate that GraphVid's semantic interaction representation enables a more faithful realization of the prompt across diverse contexts without sacrificing visual fidelity.

\FigUserStudy
\section{Qualitative Results}
\label{app:addn_qual_results}
\FigSuppQualitativeComparisonMulti
\FigSuppQualitativeComparisonSingle

Figure~\ref{fig:qualitative_multi_comp_interactions} presents qualitative comparisons of multi-object interaction control across several baseline methods and GraphVid. Given an input image and user-specified interaction cues, prior approaches often struggle to faithfully follow the intended interaction dynamics. In the ship scenario (top), baseline methods either fail to separate the motion trajectories of the two ships or introduce visual artifacts during generation. In contrast, GraphVid produces coherent diverging trajectories consistent with the user-specified interaction. In the jogging scenario (bottom), competing methods exhibit unstable motion patterns, including subjects disappearing from the frame, incorrect action generation (walking instead of jogging), or noticeable motion blur. GraphVid maintains stable subject identities and produces temporally consistent motion while correctly preserving the relative dynamics between the lead and follow joggers. These examples highlight the advantage of structured interaction graph conditioning for modeling coordinated multi-object dynamics.

Figure~\ref{fig:qualitative_single_comp_interactions} presents additional qualitative comparisons between GraphVid and prior controllable video generation methods. Given the same input image and user-specified interaction cues, baseline approaches often struggle to maintain coherent motion dynamics across time. In the handshake scenario, competing models exhibit unstable interaction behavior, including abrupt termination of the motion trajectory or blurred hand dynamics, indicating difficulty in sustaining consistent multi-entity interactions. In contrast, GraphVid produces a stable handshake sequence with smooth temporal evolution and consistent contact between the two subjects. In the metro scenario, baseline methods either terminate the motion prematurely or introduce structural hallucinations such as incorrect wheel formations, reflecting weak grounding between motion generation and object structure. GraphVid preserves object geometry while generating a steady forward motion along the intended track. These examples further demonstrate that interaction graph conditioning improves temporal consistency and structural reliability in multi-object video generation.

\FigFailureZzz
\FigGeneratedSamples

\section{Failure Case Analysis}
While GraphVid significantly improves controllable video generation, certain limitations remain in semantic granularity and object grounding. A common failure mode occurs when the VLM fails to isolate small or symbolic elements during scene graph construction. As shown in \Cref{fig:failure_semantic_granularity}, the user attempts to apply a rotational transformation to the small ``Zzz'' symbol above the sleeping bear. However, the VLM does not segment this low-salience text as an independent node. As a result, the interaction vector is incorrectly associated with the nearest high-confidence object—the bear itself. Consequently, the generation model applies a rigid-body rotation to the entire subject rather than the intended text element, producing severe geometric distortion across frames. This example highlights a limitation of current zero-shot detection pipelines for fine-grained interaction control and suggests that future systems may benefit from explicit node specification or interactive segmentation to ensure correct grounding of small semantic elements.

\section{Broader Impacts}

Controllable video generation has the potential to significantly expand the accessibility of visual content creation. By enabling users to specify scene dynamics through intuitive interaction graphs rather than complex motion annotations, GraphVid may lower the technical barrier for generating high-quality animations and simulations. This capability could benefit applications such as creative media production, virtual prototyping, educational visualization of physical phenomena, and simulation environments for robotics and embodied AI research. In particular, representing dynamics through structured interactions may also support improved interpretability in generative systems by making motion control more explicit and modular.

At the same time, advances in controllable video synthesis also raise concerns regarding misuse. The ability to generate realistic videos from minimal input could potentially contribute to the creation of deceptive or manipulated media, including misinformation or impersonation. While GraphVid focuses on structured interaction control rather than photorealistic identity synthesis, similar generative methods may still be misused if deployed irresponsibly. We encourage future work to explore safeguards such as watermarking, provenance tracking, and detection methods to mitigate harmful applications.

\newpage

\clearpage

\begin{tcolorbox}[
  breakable,
  colback=gray!3,
  colframe=black!20,
  title={Prompt for Motion Source Classification},
  fonttitle=\bfseries,
  boxrule=0.5pt,
  arc=2pt,
  left=6pt,
  right=6pt,
  top=6pt,
  bottom=6pt
]
\small
You are an Expert in Kinematics and Video Motion Analysis.

You are given:
\begin{enumerate}[itemsep=0em, topsep=0em]
    \item A short video clip
    \item A caption describing the video: ``\{caption\}''
\end{enumerate}

Your task is to isolate true, discrete object-to-object physics from camera-induced motion and ambient environmental motion.
\vspace{0.2cm}

\textbf{Motion Categories}
\begin{itemize} [itemsep=0em, topsep=0em]
    \item \texttt{discrete\_object\_physics}: Solid, distinct objects (\eg, people, cars, bats, balls) moving due to explicit physical forces, collisions, or locomotion. The motion can be easily tracked with a bounding box.
    \item \texttt{ambient\_environmental}: The primary motion is fluid, gaseous, or driven by the environment. Examples: fire burning, water flowing, waves crashing, smoke billowing, or static trees swaying in the wind.
    \item \texttt{camera\_motion}: The camera is panning, tilting, zooming, or shaking. Key indicator: the static background elements (walls, ground) shift uniformly across the screen.
    \item \texttt{ego\_motion}: First-person perspective moving through space (\eg, POV walking, driving).
    \item \texttt{static}: Little to no meaningful motion.
\end{itemize}
\vspace{0.1cm}

\textbf{Strict Evaluation Rules}
\begin{enumerate} [itemsep=0em, topsep=0em]
    \item \textbf{The Ambient Rule (CRITICAL):} If the dominant motion in the video is fire, water, smoke, or wind, you MUST classify it as \texttt{ambient\_environmental}. Scene graphs cannot easily model fluids.
    \item \textbf{Background First:} Always look at the static background elements first. If the background is shifting uniformly, classify as \texttt{camera\_motion}.
    \item \textbf{The Tracking Rule:} If a camera pans heavily to track a moving subject (blurring the background), classify as \texttt{camera\_motion} or \texttt{ego\_motion}, because pixel-level motion is dominated by the camera.
\end{enumerate}

Return \textbf{STRICT JSON only}. You must follow this exact key order to ensure proper reasoning:

\begin{verbatim}
{
  "background_and_ambient_analysis": "Briefly check for shifting 
  backgrounds or fluid/environmental motion (fire, water, wind).",
  "object_analysis": "Briefly describe what discrete, 
  solid objects are moving.",
  "primary_motion_source": "discrete_object_physics | 
  ambient_environmental | camera_motion | ego_motion | static",
  "confidence": 0.0-1.0
}
\end{verbatim}
\end{tcolorbox}
\newpage 

\begin{tcolorbox}[
  breakable,
  colback=gray!3,
  colframe=black!20,
  title={Prompt for Initial Scene Graph Generation},
  fonttitle=\bfseries,
  boxrule=0.5pt,
  arc=2pt,
  left=6pt,
  right=6pt,
  top=6pt,
  bottom=6pt
]
\small

\textbf{ROLE}

You are a Physics-Informed Vision-Language Expert. Your goal is to analyze a static image and its corresponding Object JSON to construct a \textit{Physics Scene Graph}.

\medskip
\textbf{TASK}

Generate a directed graph of physical interactions between the provided objects. Your output must be a strict JSON object containing a list of edges.

\medskip
\textbf{Interaction Rules (The Physics Engine)}

\begin{enumerate}[itemsep=0em, topsep=0em]
\item \textbf{Visible Evidence Only:} Do not hallucinate interactions.

\item \textbf{Implicit Proximity:} If objects are not touching but are spatially related (\eg, ball near bat), infer a \texttt{spatial\_proximity} or \texttt{trajectory\_threat} edge. \textbf{Do not return an empty list.}

\item \textbf{Newtonian Terminology:} Use physics-grounded terms such as Friction, Tension, Normal Force, Gravity, Torque, or Applied Force.

\item \textbf{Full Sentence Relations:} The \texttt{relation\_description} must be grammatically complete:

[\text{Spatial Preposition}]+[\text{Source Object}]+[\text{Physics Interaction}]+[\text{Target Object}]

\end{enumerate}

\medskip
\textbf{Output Schema}

Return \textbf{ONLY valid JSON}. No explanation or additional text.

\begin{verbatim}
{
  "edges": [
    {
      "source_node_id": "obj_XXX",
      "target_node_id": "obj_YYY",
      "interaction_type": "string",
      "relation_description": "string",
      "physics_attributes": {
        "force_vector_approx": "string",
        "estimated_magnitude": "string"
      }
    }
  ]
}
\end{verbatim}

\end{tcolorbox}

\newpage

\begin{tcolorbox}[
  breakable,
  colback=gray!3,
  colframe=black!20,
  title={Prompt for Synthetic Graph Update (Training)},
  fonttitle=\bfseries,
  boxrule=0.5pt,
  arc=2pt,
  left=6pt,
  right=6pt,
  top=6pt,
  bottom=6pt
]
\small

\textbf{ROLE}

You are a \textit{Synthetic User Simulator} for an interactive video generation tool.
Your goal is to translate a textual description of a video into the \textbf{minimal set of manual graph edits} a human user would perform to direct that motion.

\medskip
\textbf{TASK}

Analyze the \textbf{Initial State} versus the \textbf{Target Description} and \textbf{Detailed Video Description}.  
Determine the primary causal actions and convert them into graph operations (\texttt{Upsert}/\texttt{Remove}).  
Additionally, return a list of candidate nodes that will have dynamic motion (\ie, will move in the video).

\medskip
\textbf{Input Data}

\begin{enumerate}[itemsep=0em, topsep=0em]
\item \textbf{Initial Object List} (provided JSON)
\item \textbf{Target Description} (ground-truth caption)
\item \textbf{Detailed Video Context Description} (descriptive analysis of the video)
\end{enumerate}

\medskip
\textbf{Core Philosophy}

\begin{enumerate}[itemsep=0em, topsep=0em]
\item \textbf{Action-Driven Causality:} The scene graph acts as a control interface. Model only the primary, active drivers of motion (\eg, A pushes B). Ignore reactive, passive, or static forces unless they directly cause a state change.

\item \textbf{One-Way Causality:} Create edges strictly in the direction of the action (\eg, A pushes B $\rightarrow$ A$\rightarrow$B).

\item \textbf{Sparsity Constraint:} Limit connections to a maximum of two edges between any pair of nodes. Merge multiple interactions into a single \texttt{relation\_description}.

\item \textbf{Motion Updates Required:} Carefully analyze the context to identify physical motion. Output at least one edge modification (\texttt{upsert} or \texttt{remove}) whenever physical movement occurs.

\item \textbf{Order of Operations:} When modifying an existing relationship, issue the \texttt{remove\_edges} command for the old edge before issuing the \texttt{upsert\_edges} command for the new state.

\item \textbf{Isolate Object Physics from Camera Ego-Motion:} If object movement is solely caused by camera motion (panning, zooming, tracking), omit edges for those background objects. Only model physical object-to-object interactions.

\item \textbf{Node Consistency Constraint:} You must use \textbf{only} the exact \texttt{node\_id} strings provided in the Object List. Never invent or hallucinate new node IDs (\eg, \texttt{surface\_implied} or \texttt{ground}). If a relevant object is missing, describe the interaction using the closest available physical node.
\end{enumerate}

\medskip
[Output Schema]

\end{tcolorbox}
\newpage

\begin{tcolorbox}[
  breakable,
  colback=gray!3,
  colframe=black!20,
  title={Output Schema for Synthetic Graph Update},
  fonttitle=\bfseries,
  boxrule=0.5pt,
  arc=2pt,
  left=6pt,
  right=6pt,
  top=6pt,
  bottom=6pt
]

\textbf{Output Schema}

Return \textbf{ONLY valid JSON}.

\begin{verbatim}
{
  "observation_notes": String (Max 2 sentences. Identify the 
    active subjects and how they are supposed to move based on 
    provided context),
  "remove_nodes": [],
  "remove_edges": [
    { "source_node_id": "obj_A", "target_node_id": "obj_B" }
  ],
  "upsert_edges": [
    {
      "source_node_id": "obj_XXX",
      "target_node_id": "obj_YYY",
      "interaction_type": String (Choose ONE: push, pull, 
        hold, contact, support, ...)",
      "relation_description": String (Concise description of 
        the user's intent, e.g., `Man places left hand on 
        woman's shoulder'),
      "physics_attributes": {
        "force_vector_approx": String (e.g., downward,
        toward_camera, lateral),
        "estimated_magnitude": String (e.g., high, medium, 
          low)
      }
    }
  ]
}
\end{verbatim}

\end{tcolorbox}

\newpage

\begin{tcolorbox}[
  breakable,
  colback=gray!3,
  colframe=black!20,
  title={Prompt for Inference-Time User Intent Translation},
  fonttitle=\bfseries,
  boxrule=0.5pt,
  arc=2pt,
  left=6pt,
  right=6pt,
  top=6pt,
  bottom=6pt
]
\small

\textbf{ROLE}

You are a Physics-Informed Vision-Language Expert acting as a Scene Graph Compiler for a controllable video generation system.
Your task is to translate the user's explicit interaction intent into the strict JSON schema expected by the downstream physics engine.

\medskip
\textbf{Input Data}

\begin{enumerate}[itemsep=0em, topsep=0em]
\item \textbf{Current Graph State:} all available nodes and initial edges for the scene graph
\item \textbf{User Action Record:} user-provided interaction context, including vectors where applicable
\item \textbf{Image Evidence:} recorded user interactions for improving context understanding
\end{enumerate}

\medskip
\textbf{Rules}

\begin{enumerate}[itemsep=0em, topsep=0em]
\item \textbf{One-Way Causality:} Create edges strictly in the direction of the action (source $\rightarrow$ target).

\item \textbf{Vocabulary Constraints:} Restrict \texttt{interaction\_type} to simple structural verbs (\eg, \texttt{push}, \texttt{pull}, \texttt{hold}, \texttt{contact}, \texttt{support}, \texttt{spatial\_proximity}).

\item \textbf{Relation Enrichment:} Set \texttt{relation\_description} to the user's exact context, slightly enriched to describe the physical mechanics.

\item \textbf{Physics Deduction:} Deduce the underlying \texttt{physics\_attributes} needed to execute the motion. \texttt{force\_vector\_approx} must use simple terms (\eg, \texttt{downward}, \texttt{toward\_camera}, \texttt{lateral}, \texttt{upward}, \texttt{neutral}). \texttt{estimated\_magnitude} must be one of \texttt{high}, \texttt{medium}, or \texttt{low}.

\item \textbf{Grounding:} Never hallucinate new node IDs. Use only the IDs provided.
\end{enumerate}

\medskip
\textbf{Output Schema}

Return \textbf{ONLY valid JSON}.

\begin{verbatim}
{
  "observation_notes": String (Briefly explain how you mapped the 
  user intent to the physics attributes),
  "remove_nodes": [],
  "remove_edges": [],
  "upsert_edges": [
    {
      "source_node_id": "obj_XXX",
      "target_node_id": "obj_YYY",
      "interaction_type": "String",
      "relation_description": "String",
      "physics_attributes": {
        "force_vector_approx": "String",
        "estimated_magnitude": "String"
      }
    }
  ]
}
\end{verbatim}

\end{tcolorbox}

\begin{tcolorbox}[
  breakable,
  colback=gray!3,
  colframe=black!20,
  title={Prompt for Standard Video Captioning},
  fonttitle=\bfseries,
  boxrule=0.5pt,
  arc=2pt,
  left=6pt,
  right=6pt,
  top=6pt,
  bottom=6pt
]
\small

Please describe the video in a concise and natural paragraph. Your description should follow these rules:

\begin{enumerate}[itemsep=0em, topsep=0em]
\item \textbf{Primary Motion Focus:} Focus primarily on the motion and behavior of the main subjects in the video (\eg, people or animals). Describe their actions in chronological order.

\item \textbf{Subject Appearance:} Briefly describe the appearance and number of the main subjects, including attributes such as color, size, and orientation.

\item \textbf{Spatial Relationships:} Mention spatial relationships between subjects when relevant (\eg, in front of, to the left of, behind).

\item \textbf{Camera Perspective:} At the end of the description, specify the camera perspective and movement, including the shooting angle (\eg, top-down, frontal, side view) and camera motion (\eg, pan left, zoom in, dolly out, slight shake), particularly when they influence perceived motion.

\item \textbf{Scene Context:} Briefly describe the background or environment, but keep this minimal unless it is necessary to understand the motion.

\item \textbf{Restrictions:} Do not include text recognition, named characters, or stylistic analysis (\eg, realistic, animated) unless they are essential for understanding the motion.
\end{enumerate}

\medskip
\textbf{Output Constraint}

Provide a concise and fluent paragraph, ideally \textbf{2--5 sentences} in length.

\end{tcolorbox}